\documentclass{bmvc2k}

\title{Reachability Is Not Generalization: Understanding Verb--Noun Decomposition in Assembly Action Recognition}

\addauthor{Changyi Li}{changyi.li@aalto.fi}{1}
\addauthor{Yu Xiao}{yu.xiao@aalto.fi}{1}
\addinstitution{
 Department of Information and\\
 Communications Engineering\\
 Aalto University\\
 02150 Espoo, Finland
}

\runninghead{Li, Xiao}{Understanding Verb--Noun Decomposition}

\usepackage{amssymb}      
\usepackage{booktabs}     
\usepackage{multirow}     
\usepackage{makecell}     
\usepackage{array}
\graphicspath{{figures/}}   

\begin{document}

\maketitle

\begin{abstract}
Assembly actions are compositional: they combine a manipulation with a part or tool. In deployment, systems routinely encounter novel combinations of familiar components, yet an atomic action classifier assigns every unseen combination exactly zero probability by construction. The prevailing solution is verb--noun decomposition, which predicts components separately and recombines them to reach unseen actions. While widely adopted, how decomposition generalizes under compositional shift remains poorly understood. We present a systematic analysis of verb--noun decomposition across three assembly datasets (MECCANO, HAViD, and IMPACT). Although decomposition escapes the atomic ceiling, its generalization extends only partially beyond it. Unseen-composition performance remains strongly tied to the co-occurrence structure of the training data, indicating that much of the observed gain arises from interpolation within densely supported regions of the compositional space rather than from unconstrained recombination. Across datasets, failures consistently concentrate on the larger-vocabulary component, and IMPACT's verb-heavy vocabulary reverses the bottleneck from nouns to verbs. We further show that shared-encoder training introduces component entanglement, encouraging reliance on co-occurrence patterns that transfer poorly to unseen compositions and trailing independent recombination by up to $6.0\times$ in harmonic mean. Taken together, these findings explain why decomposition achieves only partial compositional generalization in practice. By identifying primitive support, vocabulary asymmetry, and component entanglement as connected sources of error, we provide a portable diagnostic framework for studying compositional recognition beyond aggregate accuracy. Code: https://github.com/hisalaheiyo/assembly.
\end{abstract}

\section{Introduction}
\label{sec:intro}

Assembly action recognition is an important capability for industrial AI systems, including robotic learning from demonstrations, assembly monitoring, and procedural guidance. A defining property of this setting is compositionality: actions are naturally formed by combining a manipulation (verb) with a part or tool (noun). As the verb and noun vocabularies grow, the number of possible actions increases multiplicatively, while annotated datasets cover only a sparse subset of the resulting composition space. Across the datasets studied in this work, observed verb--noun pairs occupy only $22.1\%$ to $51.7\%$ of the full $|\mathcal{V}| \times |\mathcal{N}|$ grid. In deployment, systems therefore encounter not entirely new components, but novel compositions of familiar ones.

This sparsity exposes a fundamental limitation of the traditional atomic formulation. Atomic models treat each observed verb--noun pair as a separate class and learn a classifier over a closed action vocabulary. Any unseen composition therefore receives zero probability mass. We refer to this restriction as the \emph{atomic ceiling}: a limitation imposed by the hypothesis class itself rather than by the learning algorithm. The prevailing alternative is decomposition, which predicts verbs and nouns separately and recombines them through $\log P(v)+\log P(n)$. By construction, decomposition removes the atomic ceiling and makes unseen compositions reachable. Reachability, however, is not the same as generalization. While decomposition routinely succeeds on standard train/test splits, much less is known about the regime that motivates it in the first place: unseen compositions.

Successful recombination implicitly relies on several assumptions. Components must remain sufficiently separable for independent prediction; factorized posteriors must transfer to unseen combinations; learning difficulty should not be dominated by one component; and the chosen decomposition should provide an appropriate level of abstraction. These assumptions span representation, prediction, learning dynamics, and label granularity, yet none is guaranteed in practice. Understanding which of them hold, and which fail, is essential for understanding the limits of compositional generalization.

Our goal is therefore not to ask whether decomposition outperforms atomic prediction on average. Some form of decomposition is already necessary to overcome the atomic ceiling. Instead, we ask what limits compositional generalization once that ceiling has been removed. To answer this question, we study three representative formulations: atomic prediction, which defines the unreachable baseline; the shared-encoder multi-head architecture that dominates current practice; and independent recombination of separately trained component models. We take the latter as the primary object of diagnosis: if even the strongest realization of decomposition exhibits shallow generalization to unseen compositions, the limitation belongs to the paradigm itself rather than to a particular implementation.

Our analysis proceeds from phenomenon to localization, mechanism, and intervention. We first ask whether unseen-composition success reflects genuine recombination or merely interpolation within densely observed regions of the compositional space (\textbf{RQ1}). We then examine where failures occur and whether the dominant source of error can be predicted before training (\textbf{RQ2}). Having localized the bottleneck, we investigate why the field-standard shared encoder generalizes worse than independent recombination (\textbf{RQ3}). Finally, we evaluate whether these limitations can be alleviated through alternative factorization granularities, architectures, or objectives (\textbf{RQ4}).

In summary, this paper makes three contributions. First, we provide the first systematic analysis of verb--noun decomposition under compositional shift and develop a set of diagnostics that connect observed behavior to the assumptions underlying factorized action recognition. Second, we show that unseen-composition failures are dominated by the larger-vocabulary component, revealing a predictable bottleneck that depends on dataset structure rather than on whether the component corresponds to verbs or nouns. Third, we identify component entanglement as a key source of the gap between shared and independent decomposition, explain its impact on compositional generalization, and derive practical implications for the design of compositional recognition systems.

\begin{figure}[t]
\centering
\includegraphics[width=0.8\linewidth]{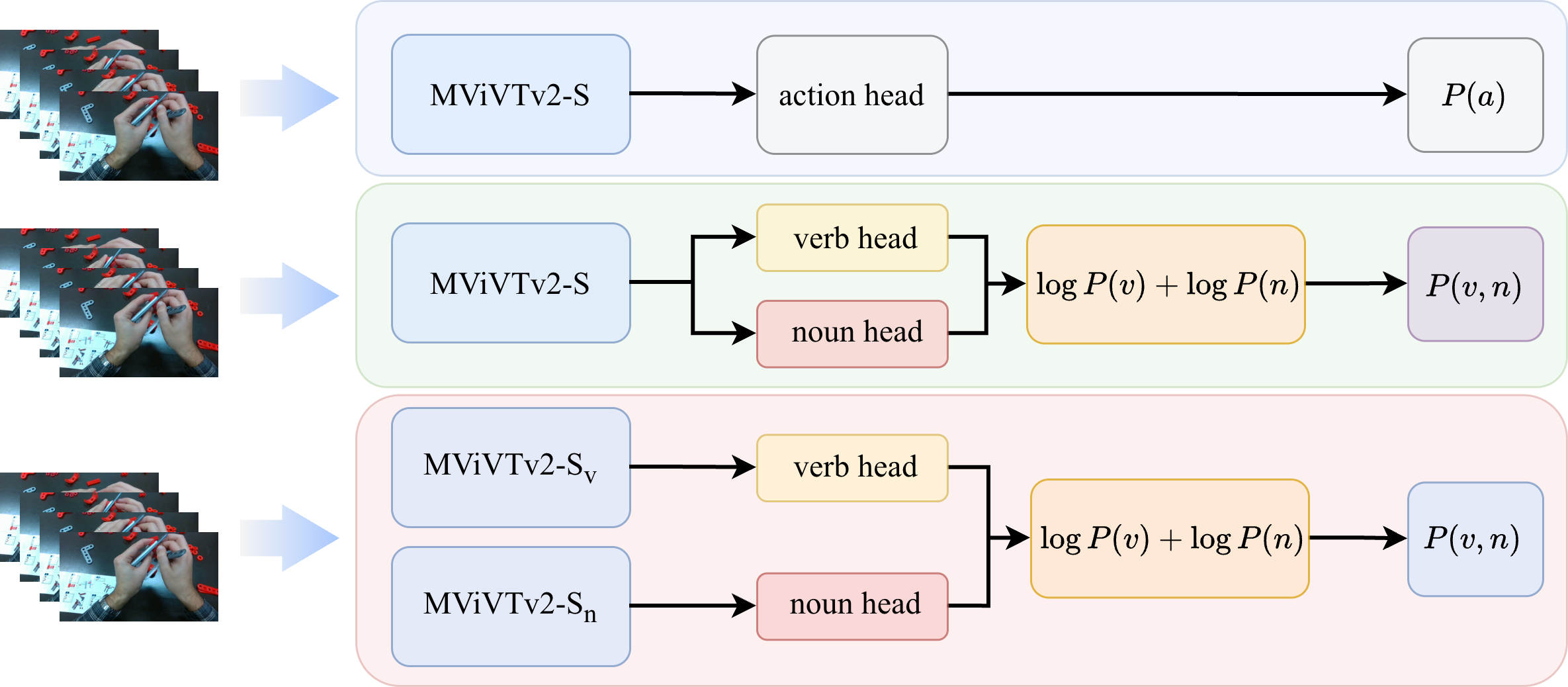}

\caption{
\textbf{Recognition formulations studied in this paper.}
Top: \emph{atomic prediction}; middle: \emph{shared decomposition}; bottom: \emph{independent decomposition}. Atomic prediction treats each action as a single class, while the decomposition variants predict verbs and nouns separately using either a shared or independent backbone and combine them via $\log P(v)+\log P(n)$. The coloured output boxes indicate the reachable output space: atomic prediction is limited to training actions, whereas decomposition spans the full verb$\times$noun space and can represent unseen compositions.
}

\label{fig:framework}
\end{figure}

\section{Related Work}
\label{sec:related}

Our work lies at the intersection of assembly action recognition, compositional generalization, and diagnostic analyses of learned recognition systems. Existing assembly-recognition methods largely adopt verb--noun decomposition as a practical solution to combinatorial action spaces; compositional-learning methods seek to improve performance within that formulation; and diagnostic studies provide tools for examining the assumptions underlying successful modelling choices. Rather than proposing a new recognition architecture or compositional objective, we study decomposition itself and ask when, why, and to what extent it supports compositional generalization.

\paragraph{Assembly recognition and verb--noun decomposition.}
Assembly and procedural understanding benchmarks span industrial assembly, instructional videos, and egocentric perception~\cite{coin2019,procel2023,meccano2021,havid2022,assembly101,impact2026,epic100,ego4d,egoexo4d, li2025revisiting}. Despite substantial differences in viewpoint, interaction complexity, and annotation granularity, most action-recognition systems adopt a common factorization of actions into operations and manipulated objects~\cite{tbn,rulstm,videomae,mvitv2,timesformer,vivit}. The motivation is straightforward: decomposition provides a scalable way to address the combinatorial growth of action spaces. Yet decomposition is typically evaluated through downstream recognition accuracy rather than through direct examination of the conditions that justify it. Our work focuses on those conditions directly.

\paragraph{Compositional generalization.}
A large body of work studies recognition of unseen combinations of known primitives. In images, compositional zero-shot learning investigates unseen attribute--object pairs using shared primitive representations~\cite{czsl_misra,symnet,cge,decomposed,causal_comp,csp,troika,compcos}. In video, compositional action recognition evaluates held-out verb--object pairs and develops objectives that encourage compositional representations~\cite{somethingelse,c2c}. While these methods differ substantially in architecture and supervision, they share a common assumption: decomposition provides a suitable substrate for generalizing to unseen compositions. Our work is complementary. Rather than introducing a new compositional objective, we examine the assumptions that existing methods implicitly rely upon and use C2C only as a representative intervention within that analysis.

\paragraph{Diagnostic analyses of recognition systems.}
A complementary line of research treats trained models as objects of scientific investigation rather than solely as benchmark competitors. Studies of shortcut learning and representation analysis have shown that strong performance can arise from spurious correlations and have examined what visual models actually encode~\cite{shortcut,kowal_deeper,sigurdsson_actions,texture_bias,scene_bias}. The common goal is to expose the assumptions underlying successful modelling choices and measure the extent to which those assumptions hold in practice. We adopt the same perspective for verb--noun decomposition, treating its underlying conditions as testable hypotheses and evaluating when, why, and to what extent they support compositional generalization under unseen-composition shift.

In summary, existing work either adopts decomposition as a practical modelling choice or seeks to improve performance within it. We instead study decomposition itself, asking what governs its success and failure under compositional shift.

\section{Testing the Assumptions of Decomposition}
\label{sec:framework}

Verb--noun decomposition removes the atomic ceiling and makes unseen compositions reachable. Whether those compositions can also be recognized correctly depends on a set of implicit assumptions. This section formalizes those assumptions and introduces the diagnostics used to test them.

Figure~\ref{fig:framework} summarizes the three recognition formulations studied in this work. Atomic prediction serves as the unreachable baseline, shared decomposition represents the dominant formulation in current practice, and independent decomposition provides the cleanest realization of the factorization assumption. We use these formulations as reference points throughout the paper to isolate the sources of compositional success and failure.

\subsection{Notation and Recognition Formulations}
\label{sec:framework:notation}

A clip $x$ contains one labelled action segment. Let $\mathcal{V}$ and $\mathcal{N}$ denote the verb and noun vocabularies. The action vocabulary $\mathcal{A}\subseteq\mathcal{V}\times\mathcal{N}$ contains the verb--noun compositions that occur at least once, with $|\mathcal{A}|\ll|\mathcal{V}|\cdot|\mathcal{N}|$. Each action label is written as $a=(v_a,n_a)$, where $v_a$ and $n_a$ denote the verb and noun components of action $a$.

We consider two evaluation regimes. The \emph{standard split} samples train and test segments i.i.d.\ over $\mathcal{A}$. The \emph{compositional split} partitions actions into disjoint sets $\mathcal{A}_\text{seen},\mathcal{A}_\text{unseen}$, while ensuring that every primitive appearing in $\mathcal{A}_\text{unseen}$ is observed during training in other compositions. This setting isolates the ability to recombine familiar components into novel actions.

Atomic prediction treats each observed action as a separate class and places a classifier over $\mathcal{A}_\text{train}$. Under the compositional split,

\begin{equation}
\mathrm{Acc}^{\text{top-}K}_\text{atom}(\mathcal{A}_\text{unseen}) = 0 \quad\text{for all } K \le |\mathcal{A}_\text{train}|,
\label{eq:atomic_zero}
\end{equation}

which defines the atomic ceiling: unseen compositions are outside the model's output space by construction.

Decomposition removes this restriction by predicting components separately. For an action $a=(v_a,n_a)$, the decomposed score is

\begin{equation}
\log \hat{P}_\text{dec}(a{=}(v_a,n_a)\mid x) = \log \hat{P}(v_a\mid x) + \log \hat{P}(n_a\mid x),
\label{eq:recombine}
\end{equation}

Equation~\eqref{eq:recombine} is defined for every $(v,n)\in\mathcal{V}\times\mathcal{N}$, so the ceiling in Eq.~\eqref{eq:atomic_zero} disappears. Whether the resulting scores remain meaningful for unseen compositions is the empirical question studied throughout the remainder of the paper.

We analyze decomposition along three design axes. The first concerns the encoder architecture: a \emph{shared} model predicts verbs and nouns from a common backbone, whereas \emph{independent} recombination trains one backbone per component and combines predictions only through Eq.~\eqref{eq:recombine}. The second concerns granularity, ranging from atomic prediction ($k{=}1$) to HAViD's four primitives $(v,m,t,l)$. The third concerns the learning objective, where we optionally add C2C~\cite{c2c} as a compositional intervention. Because independent recombination most closely matches the intended factorization, it serves as the primary reference point in our analysis.

\subsection{What Must Hold for Decomposition to Generalize?}
\label{sec:framework:assumptions}

Decomposition succeeds only if several conditions hold. Each condition motivates a corresponding diagnostic, allowing us to attribute failures to specific properties of the formulation rather than treating decomposition as a monolithic design choice.

\paragraph{Visual separability.}
Decomposition assumes that the visual evidence required to recognize one component can be represented without being inseparably tied to the other. In the ideal case, the score assigned to a verb should remain stable across different noun contexts:

\begin{equation}
\hat{P}(v\mid x_{v,n_1}) \approx \hat{P}(v\mid x_{v,n_2}) \quad \forall\, n_1,n_2 .
\label{eq:A1}
\end{equation}

When this condition fails, a component head may rely on its partner as a shortcut rather than learning the intended primitive. To evaluate this possibility, we compare shared and independent decomposition. Each component is trained once jointly with its partner and once in isolation, and we measure the additional seen-to-unseen degradation introduced by joint training:

\begin{equation}
\Delta_\text{ent} = \mathrm{drop}^{\text{joint}}_{\text{seen}\to\text{unseen}} - \mathrm{drop}^{\text{solo}}_{\text{seen}\to\text{unseen}} .
\label{eq:D1}
\end{equation}

A near-zero seen margin together with a large positive $\Delta_\text{ent}$ indicates that shared representations have introduced a co-occurrence shortcut that does not transfer to unseen compositions.

\paragraph{Statistical factorization.}
The recombination rule in Eq.~\eqref{eq:recombine} is meaningful only if component predictions provide a useful approximation to the joint action distribution:

\begin{equation}
P(v,n\mid x) \approx \hat{P}(v\mid x)\,\hat{P}(n\mid x),
\label{eq:A2}
\end{equation}

A factorized model should therefore generalize by recombining primitives rather than merely interpolating within densely observed regions of the training grid. We test this distinction directly. For each unseen composition $(v_u,n_u)$, we compare its accuracy with the amount of training support available for its constituent primitives:

\begin{equation}
\rho = \mathrm{corr}\bigl(\mathrm{Acc}_{(v_u,n_u)},\; f(v_u)\cdot f(n_u)\bigr),
\label{eq:D2}
\end{equation}

where $f(\cdot)$ counts training segments containing that primitive in any other composition. A high $\rho$ suggests that unseen accuracy is largely driven by interpolation near densely observed regions of the compositional grid, whereas a weak correlation provides stronger evidence for genuine recombination.

\paragraph{Component asymmetry.}
Assembly datasets are rarely balanced. Verb and noun vocabularies often differ substantially in size, frequency, and visual variability. As a result, one component may be inherently more difficult to learn than the other. This motivates a testable possibility: unseen-composition failures may concentrate disproportionately on the larger-vocabulary component. The hypothesis is dataset-level and can be assessed before training; IMPACT, whose verb vocabulary exceeds its noun vocabulary, provides a natural reversal of the more common noun-heavy setting.

We evaluate this possibility through a per-component error decomposition. Each unseen prediction falls into one of four cases: both components correct, verb correct but noun wrong, noun correct but verb wrong, or both wrong. Let $\pi_v$ denote the fraction of cases where the verb is correct but the noun is wrong, and $\pi_n$ the fraction where the noun is correct but the verb is wrong. Symmetric difficulty predicts $\pi_v\!\approx\!\pi_n$, whereas systematic asymmetry predicts dominance by one component.

\paragraph{Granularity.}
The previous conditions concern whether a given decomposition can generalize. Granularity asks a different question: whether the chosen decomposition is the right one in the first place. The standard verb--noun split may adequately capture the structure of some datasets, whereas richer annotation schemes such as HAViD's four-way decomposition may expose structure that a two-factor representation obscures.

We therefore evaluate whether the standard two-way decomposition is Pareto-optimal in the number of heads $k$:

\begin{equation}
\nexists\, k\neq 2:\ \mathrm{Acc}^{(k)}_\text{seen}\!\ge\!\mathrm{Acc}^{(2)}_\text{seen}\ \wedge\ \mathrm{Acc}^{(k)}_\text{unseen}\!\ge\!\mathrm{Acc}^{(2)}_\text{unseen}.
\label{eq:A4}
\end{equation}

We test this by placing each granularity at its
$(\mathrm{Acc}^{(k)}_\text{seen},\mathrm{Acc}^{(k)}_\text{unseen})$
point and reading the Pareto frontier. On HAViD we evaluate $k{=}1\ldots4$; on MECCANO and IMPACT we compare the available $k\!\in\!\{1,2\}$ settings.

Together, these diagnostics turn decomposition from a modelling convention into an analyzable object. Rather than asking only whether decomposition works, they allow us to ask why it succeeds, where it fails, and which dataset properties predict those failures. The remainder of the paper applies these diagnostics across MECCANO, HAViD, and IMPACT to characterize the behavior of decomposition under compositional shift.

\section{Experimental Protocol}
\label{sec:datasets}

\subsection{Datasets and Splits}

We evaluate decomposition on three assembly datasets whose differences allow us to distinguish paradigm-level effects from dataset-specific artifacts (Table~\ref{tab:datasets}).

\textbf{MECCANO}~\cite{meccano2021} is an egocentric toy-assembly dataset with 12 verbs and 23 objects. Its compact vocabulary and single-handed interactions make it the most controlled setting in our study. \textbf{HAViD}~\cite{havid2022} captures a bimanual industrial wiring task from three synchronized cameras and provides four primitive labels (verb, manipulated object, target object, tool; $11/29/25/5$ classes). It is the only dataset that supports evaluating alternative factorization granularities. We use all six (view, hand) subtasks. \textbf{IMPACT}~\cite{impact2026} contains five synchronized views of 112 assembly recordings. Unlike MECCANO and HAViD, it contains more verbs (22) than objects (18), providing a natural test of the component-asymmetry hypothesis.

\begin{table}[t]\centering
\caption{
\textbf{Structural diversity of the datasets used in this study.}
IMPACT uniquely reverses the common noun-heavy regime, enabling a direct test of whether the bottleneck follows vocabulary size rather than component type.
}
\label{tab:datasets}\small\setlength{\tabcolsep}{4pt}
\begin{tabular}{l ccc c c l}
\toprule
\textbf{Dataset} & $|\mathcal{V}|$ & $|\mathcal{N}|$ & grid & obs.\ (sparsity) & larger vocab & viewpoints / hands \\
\midrule
MECCANO & 12 & 23 & 276 & 61 (22.1\%) & noun ($1.9\times$) & egocentric, single-hand \\
HAViD   & 11 & 29 & 319 & 165 (51.7\%) & object ($2.6\times$) & 3 static, bimanual \\
IMPACT  & 22 & 18 & 396 & 136 (34.3\%) & \textbf{verb} ($1.2\times$) & 5 views (incl.\ ego), bimanual \\
\bottomrule
\end{tabular}\end{table}


Each dataset is evaluated under two regimes. The \emph{standard split} follows the official partitions of MECCANO and HAViD; for IMPACT, we construct a recording-level, hand-stratified split using iterative multilabel stratification~\cite{sechidis} (approximately $70/15/15\%$). Under this regime, all test actions are observed during training.

The \emph{compositional split} evaluates unseen verb--object combinations. Following prior work~\cite{somethingelse,c2c}, we remove a subset of verb--object pairs from training while ensuring that every held-out verb and object remains observable through other pairings. We hold out 6 pairs on MECCANO, 13 on HAViD, and 20 pairs per hand on IMPACT. Exact pair lists and statistics are provided in the supplementary material.

\subsection{Recognition Formulations and Training}

All models use an MViTv2-S backbone~\cite{mvitv2} pretrained on Kinetics-400 and fine-tuned end-to-end. We evaluate the three formulations of Section~\ref{sec:framework}: atomic prediction, shared decomposition, and independent decomposition. Shared and atomic models use a single encoder, differing only in their prediction heads, whereas independent decomposition trains one encoder per component and combines predictions through Eq.~\eqref{eq:recombine}.

To test whether compositional objectives provide benefits beyond decomposition itself, we additionally evaluate C2C~\cite{c2c} on top of both shared and independent formulations. All models are trained on $T{=}16$ RGB frames at $224^2$ resolution using AdamW and cosine learning-rate decay. Full hyperparameters and implementation details are reported in the supplementary material.

Following the original protocol, C2C-shared is trained end-to-end, whereas C2C-indep freezes the pretrained component encoders and trains only the composition module. Atomic models and independent component heads are selected using seen-only validation metrics; shared, granularity, and C2C models are selected using the harmonic mean of seen and unseen accuracy~\cite{c2c,cge,compcos}.

\subsection{Evaluation}

We report action top-$K$ accuracy ($K\in{1,3,5,10}$), computed either from the atomic classifier or from the recombined score of Eq.~\eqref{eq:recombine}. We additionally report per-component verb and object top-$K$ accuracy. For HAViD, joint top-$K$ requires all predicted components to appear within their respective top-$K$ sets.

Because seen and unseen performance often trade off, compositional results are summarized using the harmonic mean $\mathrm{HM}=2\,\mathrm{Acc}_\text{seen}\mathrm{Acc}_\text{unseen}/(\mathrm{Acc}_\text{seen}+\mathrm{Acc}_\text{unseen})$. Action accuracy and HM measure overall performance, while the diagnostics of Section~\ref{sec:framework} identify the sources of success and failure. For HAViD, standard-split accuracy is taken from the published MViTv2-S benchmark~\cite{havid2022}; all compositional-split results are our measurements.

\section{Diagnosing Decomposition}
\label{sec:results}

We begin with the compositional-split results, which establish the empirical
basis for the following analyses. The key question is not whether decomposition
expands the reachable action space, but why only part of it is realized. The
four research questions examine this gap through coverage dependence, component
asymmetry, encoder entanglement, and factorization granularity.

\subsection{What the Compositional Split Reveals}
\label{sec:results:anchor}

\begin{table}[t]\centering
\caption{
\textbf{Performance on the compositional split.}
Seen and unseen action top-1 accuracy and harmonic mean (HM) for five recognition formulations across three datasets. Atomic prediction serves as the unreachable baseline, while shared and independent decomposition make unseen compositions reachable. This table provides the reference point for all subsequent analyses.
}
\label{tab:master}\small
\renewcommand{\arraystretch}{0.7}
\begin{tabular}{l l ccc}
\toprule
\textbf{Dataset} & \textbf{Config} & \textbf{Seen} & \textbf{Unseen} & \textbf{HM} \\
\midrule
\multirow{5}{*}{MECCANO}
  & Atomic & 48.9 & 0.0 & 0.0 \\
  & Shared & 48.4 & 1.7 & 3.3 \\
  & Independent & 44.3 & 12.8 & 19.8 \\
  & C2C-shared & 27.9 & 13.2 & 17.9 \\
  & C2C-indep & 46.6 & 16.5 & 24.3 \\
\midrule
\multirow{5}{*}{HAViD}
  & Atomic & 54.7 & 0.0 & 0.0 \\
  & Shared & 51.1 & 7.9 & 13.7 \\
  & Independent & 52.8 & 21.1 & 30.2 \\
  & C2C-shared & 52.7 & 19.9 & 28.9 \\
  & C2C-indep & 52.0 & 21.9 & 30.7 \\
\midrule
\multirow{5}{*}{IMPACT}
  & Atomic & 52.9 & 0.0 & 0.0 \\
  & Shared & 51.0 & 6.1 & 10.6 \\
  & Independent & 49.0 & 23.3 & 31.2 \\
  & C2C-shared & 47.9 & 14.7 & 22.1 \\
  & C2C-indep & 49.5 & 22.9 & 30.9 \\
\bottomrule
\end{tabular}\end{table}

Table~\ref{tab:master} summarizes compositional-split performance across all datasets and recognition formulations and highlights three recurring patterns.

First, standard benchmarks largely obscure the issue. Replacing atomic prediction with decomposition changes standard-split accuracy only modestly, and often improves it, suggesting little reason to question the prevailing formulation when evaluation is restricted to observed actions.

Second, decomposition is necessary but not sufficient. Atomic prediction assigns zero probability to unseen compositions by construction, whereas all decomposition variants can reach them. Yet reachability translates into only partial generalization: even the strongest formulation leaves a substantial gap between seen and unseen performance.

Third, the choice of formulation matters. Independent decomposition consistently outperforms the field-standard shared formulation, increasing harmonic mean from $3.3$ to $19.8$ on MECCANO, from $13.7$ to $30.2$ on HAViD, and from $10.6$ to $31.2$ on IMPACT. Nevertheless, large seen--unseen gaps remain (e.g., $49.0$ versus $23.3$ on IMPACT), indicating that much of the reachable compositional space is still unrealized.

The remainder of the paper explains this gap. We first ask whether unseen success reflects genuine composition or interpolation within densely observed regions of the composition grid (RQ1). We then examine where failures concentrate (RQ2), why shared encoders underperform independent recombination (RQ3), and whether alternative factorization strategies can alleviate these limitations (RQ4).

\subsection{Composition or Interpolation? (RQ1)}
\label{sec:results:rq1}

Independent decomposition recovers unseen accuracy that atomic prediction cannot achieve, but such success may reflect either genuine recombination or interpolation from densely observed training coverage. To distinguish the two, we correlate the accuracy of each unseen pair with the training support of its constituent primitives, measured by $f(v)\!\cdot\!f(n)$ (Eq.~\eqref{eq:D2}). If decomposition generalizes independently of coverage, the relationship should be weak; if success is driven by support, accuracy should increase with primitive frequency.

\begin{figure}[t]
\centering
\includegraphics[width=0.8\linewidth]{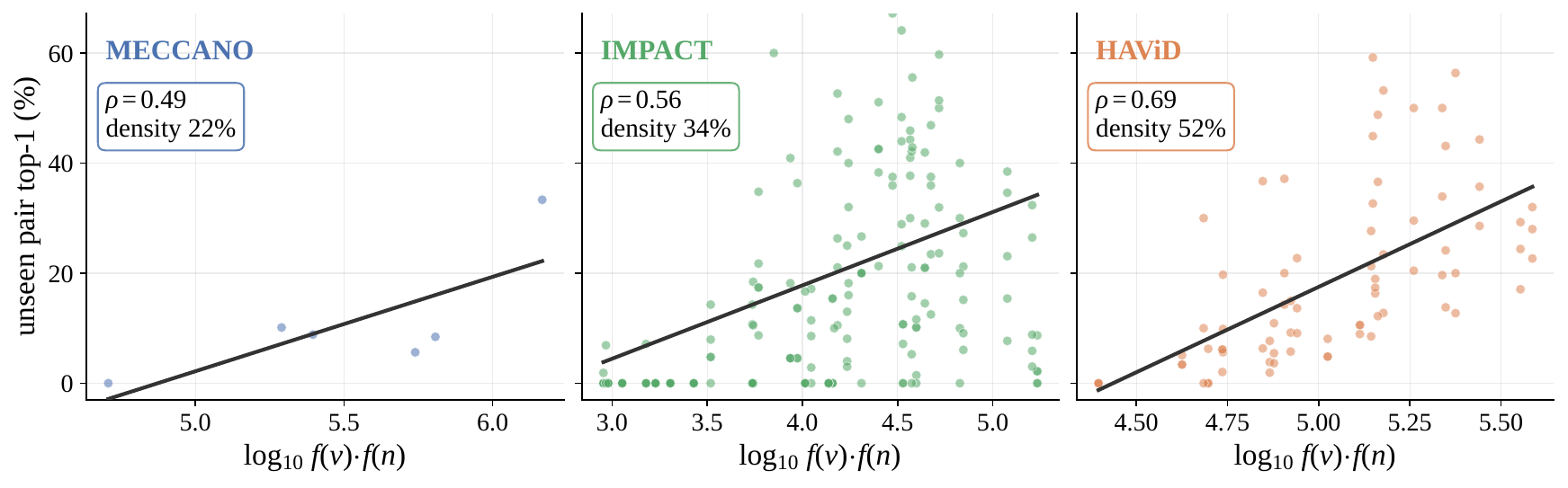}
\caption{
\textbf{Unseen accuracy increasingly follows primitive support as the composition grid becomes denser.}
Each point is an unseen verb--noun pair. Correlation strength rises from MECCANO to IMPACT to HAViD, mirroring the density of the observed composition grid. High-support pairs achieve the strongest performance, whereas low-support pairs remain near zero.
}
\label{fig:density}
\end{figure}

The correlation (Spearman's rank $\rho$) is positive on all three datasets and strengthens with composition density: $\rho=0.49$ on MECCANO, $0.56$ on IMPACT, and $0.69$ on HAViD, corresponding to grid densities of $22.1\%$, $34.3\%$, and $51.7\%$. Figure~\ref{fig:density} shows the same pattern qualitatively: unseen pairs with substantial primitive support achieve the highest accuracies, whereas low-support pairs remain near zero. Permutation tests confirm the effect ($p{=}5{\times}10^{-5}$ on HAViD and IMPACT; MECCANO alone is underpowered with six held-out pairs), and two reconstructed MECCANO splits spanning a wider support range reproduce it with $\rho{=}0.94/0.93$ (Supp.\ Secs.~13--14).

The dependence is substantial without being complete ($\rho^2=0.24$--$0.48$): decomposition extends recognition beyond the atomic ceiling, but success remains strongly conditioned on primitive support, and the remaining challenge lies in the low-support region of the grid.

This observation motivates the next question. If unseen performance depends strongly on primitive support, failures should concentrate on whichever component is hardest to cover. We therefore turn to where those failures occur and whether the dominant bottleneck can be predicted before training.

\subsection{Which Component Becomes the Bottleneck? (RQ2)}
\label{sec:results:rq2}

RQ1 showed that unseen performance depends strongly on primitive support. If so, failures should concentrate on whichever component is hardest to cover. Because larger vocabularies spread observations across more classes, we ask whether unseen errors systematically concentrate on the larger-vocabulary component.

\begin{figure}[t]
\centering
\includegraphics[width=0.8\linewidth]{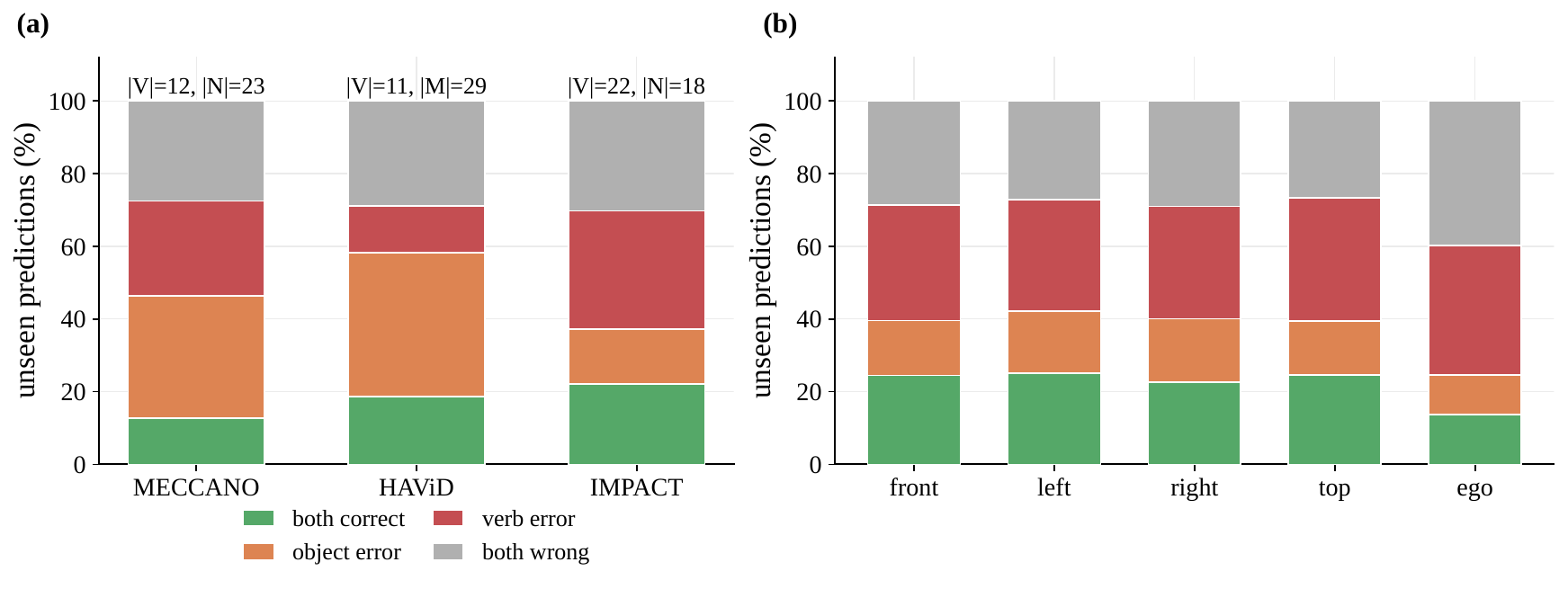}
\caption{
\textbf{The bottleneck follows vocabulary size rather than component identity.}
(a) Unseen errors partitioned by component. Failures consistently concentrate on the larger-vocabulary component. (b) The verb bottleneck in IMPACT persists across all five views, indicating that the asymmetry follows vocabulary structure rather than viewpoint.
}
\label{fig:bottleneck}
\end{figure}

To localize failures, we partition each unseen prediction according to whether the verb, object, both, or neither component is correct. If verbs and objects are equally difficult, verb-only and object-only errors should occur at similar rates. Any systematic imbalance identifies the bottleneck component. Figure~\ref{fig:bottleneck} and Table~\ref{tab:bottleneck} summarize the resulting error decomposition.

\begin{table}[t]\centering
\caption{
\textbf{The bottleneck follows vocabulary size rather than component identity.}
Across datasets, unseen failures consistently concentrate on the larger-vocabulary component. IMPACT reverses the common noun-heavy regime, and the bottleneck reverses with it across all five views.
}
\label{tab:bottleneck}\small\setlength{\tabcolsep}{4.5pt}
\renewcommand{\arraystretch}{0.7}
\begin{tabular}{l cc cccc c}
\toprule
\textbf{Dataset} & $|\mathcal{V}|$ & $|\mathcal{N}|$ & both & \makecell{obj\\err ($\pi_v$)} & \makecell{verb\\err ($\pi_n$)} & \makecell{both\\wrong} & bottleneck \\
\midrule
MECCANO & 12 & 23 & 12.8 & 33.5 & 26.2 & 27.5 & noun/obj \\
HAViD & 11 & 29 & 18.6 & 39.6 & 12.9 & 28.9 & noun/obj \\
IMPACT & 22 & 18 & 22.1 & 15.1 & 32.6 & 30.2 & verb \\
\midrule
\multicolumn{8}{l}{\footnotesize IMPACT per view (verb bottleneck is view-invariant):}\\
\quad front & & & 24.5 & 15.1 & 31.8 & 28.6 & verb \\
\quad left & & & 25.1 & 17.1 & 30.8 & 27.1 & verb \\
\quad right & & & 22.5 & 17.5 & 31.0 & 29.0 & verb \\
\quad top & & & 24.6 & 14.8 & 34.0 & 26.7 & verb \\
\quad ego & & & 13.7 & 10.9 & 35.7 & 39.7 & verb \\
\bottomrule
\multicolumn{8}{l}{\footnotesize $\pi_v$ (verb correct, object wrong) $\Rightarrow$ object fails; $\pi_n$ $\Rightarrow$ verb fails. Bottleneck $=$ larger-vocab primitive.}\\
\end{tabular}\end{table}

The asymmetry is clear and consistently follows vocabulary size. On MECCANO and HAViD, where object vocabularies substantially exceed verb vocabularies ($23$ vs.\ $12$ and $29$ vs.\ $11$), object errors dominate unseen failures ($\pi_v=33.5$ vs.\ $\pi_n=26.2$ on MECCANO; $39.6$ vs.\ $12.9$ on HAViD). IMPACT reverses the vocabulary balance ($22$ verbs vs.\ $18$ objects), and the bottleneck reverses with it: verb errors dominate ($\pi_n=32.6$ vs.\ $\pi_v=15.1$).

The bottleneck therefore follows vocabulary size rather than component identity. IMPACT provides a natural flip test, and the asymmetry flips accordingly. The effect is stable across all five IMPACT views and appears in both independent and shared formulations, indicating that it arises from the factorized representation itself rather than from viewpoint or architecture.

A practical consequence is that the dominant bottleneck can often be predicted before training from vocabulary statistics alone. Capacity, data collection, and augmentation may therefore be allocated asymmetrically toward the larger-vocabulary component. Having identified where failures concentrate, we next ask why the shared formulation amplifies them more strongly than independent recombination.

\subsection{Why Shared Training Underperforms (RQ3)}
\label{sec:results:rq3}

RQ2 identified where failures concentrate; RQ3 asks why the shared formulation amplifies them. A natural explanation is capacity: independent recombination uses one backbone per component, whereas the shared formulation relies on a single encoder. An alternative explanation is entanglement, whereby joint training encourages reliance on verb--object co-occurrence patterns that do not transfer to unseen compositions.

To distinguish between these explanations, we compare each component head when trained jointly and independently. Table~\ref{tab:sep} reports the seen margin (joint minus independent accuracy on seen compositions) and the excess unseen drop $\Delta_{\text{ent}}$ (Eq.~\eqref{eq:D1}), which measures the additional performance lost under compositional shift when the component is trained jointly. The two measurements point in opposite directions. The seen margin remains close to zero for every component, ranging from $-1.9$ to $+1.8$ pp, indicating that joint and independent training fit observed compositions equally well. Under compositional shift, however, the excess unseen drop is consistently large, reaching $6.4$ and $11.5$ pp on MECCANO, $5.9$ and $8.7$ pp on HAViD, and $12.5$ and $14.5$ pp on IMPACT.

\begin{table}[t]\centering
\caption{
\textbf{Evidence for component entanglement in shared training.}
The shortcut score $\Delta$ measures the excess seen-to-unseen drop of a jointly trained component head relative to its independently trained counterpart. Large $\Delta$ values together with near-zero seen margins indicate that shared training provides little benefit on seen compositions while substantially increasing the unseen generalization gap.
}
\label{tab:sep}\small\setlength{\tabcolsep}{5pt}
\renewcommand{\arraystretch}{0.7}
\begin{tabular}{l cc c c c}
\toprule
& \multicolumn{2}{c}{shortcut $\Delta$ (pp)} & & grid & seen-margin \\
\cmidrule(lr){2-3}
\textbf{Dataset} & verb & object & $\rho$ (D2) & density & (v, o) pp \\
\midrule
MECCANO & 6.4 & 11.5 & 0.49 & 22.1\% & (-0.3, -0.6) \\
HAViD & 5.9 & 8.7 & 0.69 & 51.7\% & (-1.2, -1.9) \\
IMPACT & 12.5 & 14.5 & 0.56 & 34.3\% & (+1.8, +0.7) \\
\bottomrule
\multicolumn{6}{l}{\footnotesize $\Delta=$ excess seen$\to$unseen drop of joint over solo (shortcut). seen-margin $\approx 0$ rules out capacity.}\\
\end{tabular}\end{table}

This pattern is difficult to reconcile with a pure capacity explanation: if the shared formulation lacked capacity, the deficit should already appear on seen compositions, yet the penalty emerges primarily on unseen ones. A matched-capacity control confirms this: freezing half of each independent tower to match the shared model's trainable parameters preserves most of the unseen advantage (Supp.\ Sec.~15). The results are instead consistent with component entanglement: shared training encourages reliance on co-occurrence structure that benefits familiar combinations but transfers poorly to novel ones. The effect is strongest on the larger-vocabulary component identified in RQ2, suggesting that entanglement amplifies the same coverage limitations observed in RQ1 and RQ2. The shared--independent gap of Table~\ref{tab:master} therefore reflects representation quality rather than model size; independent recombination generalizes better because it preserves component separation. This motivates the final question: can the resulting limitation be alleviated through architectural choices, factorization granularity, or training objectives?

\subsection{Can the Limitation Be Engineered Away? (RQ4)}
\label{sec:results:rq4}

RQ3 identified component entanglement as a major source of compositional failure. We now ask whether standard design choices can mitigate this limitation. We evaluate three common interventions: factorization granularity, architecture, and training objective. Across all three, the same pattern emerges: performance levels change, but the failure structure identified in RQ1--RQ3 remains largely intact.

\begin{figure}[t]
\centering
\includegraphics[width=0.8\linewidth]{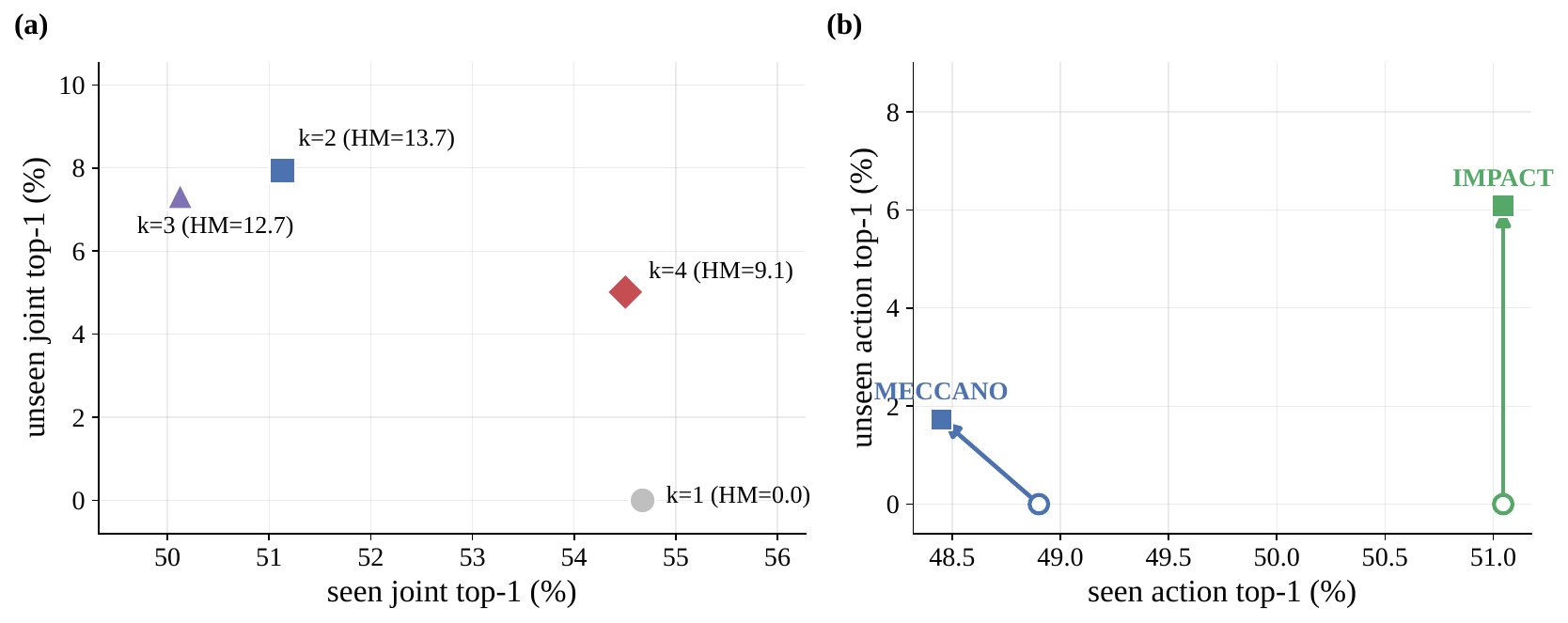}
\caption{
\textbf{More factorization does not imply better compositional generalization.}
HAViD performance under $k=1\ldots4$ factorization levels. The standard two-component decomposition ($k=2$) achieves the best unseen performance. Further factorization recovers seen accuracy but degrades generalization. Inset: the corresponding $k{=}1\!\to\!2$ transition on MECCANO and IMPACT.
}
\label{fig:granularity}
\end{figure}

We first vary factorization granularity while keeping the action label fixed. Figure~\ref{fig:granularity} and Table~\ref{tab:granularity} compare atomic prediction ($k=1$), standard verb--object decomposition ($k=2$), and finer factorizations with auxiliary target and tool heads ($k=3,4$). The standard two-component decomposition achieves the strongest unseen performance. Increasing granularity beyond $k=2$ largely recovers seen accuracy while reducing unseen performance, indicating that additional factorization improves the fit to observed compositions without improving recombination on unseen ones. Among the interventions, architecture has the largest effect. Independent recombination consistently outperforms the shared formulation, supporting the entanglement mechanism identified in RQ3. C2C mainly benefits the shared architecture, narrowing the shared--independent gap, but adds little once the components are trained independently. Thus, C2C primarily reduces the entanglement penalty rather than improving beyond architectural de-entanglement.

\begin{figure}[t]
\centering
\includegraphics[width=0.55\linewidth]{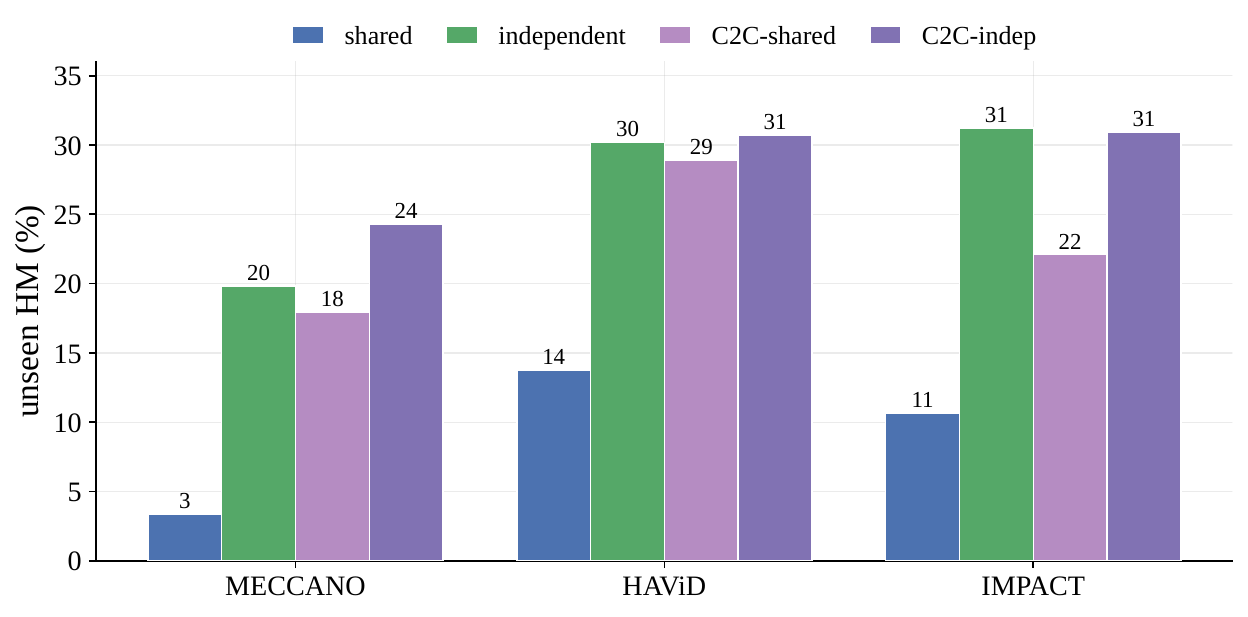}
\caption{
\textbf{Interventions change performance levels but not the bottleneck.}
Unseen HM for shared and independent formulations, with and without C2C. Independent recombination provides the largest gain. C2C substantially improves the shared baseline but adds little once components are already trained independently.
}
\label{fig:interventions}
\end{figure}

\begin{table}[t]\centering
\caption{
\textbf{Finer factorization does not improve compositional generalization.}
The standard two-component decomposition ($k=2$) provides the best unseen performance. Increasing granularity mainly improves seen accuracy while degrading transfer to unseen compositions.
}
\label{tab:granularity}\small\setlength{\tabcolsep}{5pt}
\renewcommand{\arraystretch}{0.7}
\begin{tabular}{l l cc c}
\toprule
\textbf{Dataset} & \textbf{$k$ (heads)} & \textbf{Seen} & \textbf{Unseen} & \textbf{HM} \\
\midrule
\multirow{4}{*}{HAViD}
  & $k=1$ (atomic) & 54.7 & 0.0 & 0.0 \\
  & $k=2$ ($v,m$) & 51.1 & 7.9 & 13.7 \\
  & $k=3$ ($+t$) & 50.1 & 7.3 & 12.7 \\
  & $k=4$ ($+t,l$) & 54.5 & 5.0 & 9.1 \\
\midrule
\multirow{2}{*}{MECCANO}
  & $k=1$ (atomic) & 48.9 & 0.0 & 0.0 \\
  & $k=2$ ($v,n$) & 48.4 & 1.7 & 3.3 \\
\midrule
\multirow{2}{*}{IMPACT}
  & $k=1$ (atomic) & 52.9 & 0.0 & 0.0 \\
  & $k=2$ ($v,n$) & 51.0 & 6.1 & 10.6 \\
\bottomrule
\multicolumn{5}{l}{\footnotesize HAViD action $=(v,m)$; $k{=}3,4$ add target/tool as auxiliary heads. $k{=}2$ is the unseen sweet spot.}\\
\end{tabular}\end{table}

\begin{figure}[t]
\centering
\includegraphics[width=0.8\linewidth]{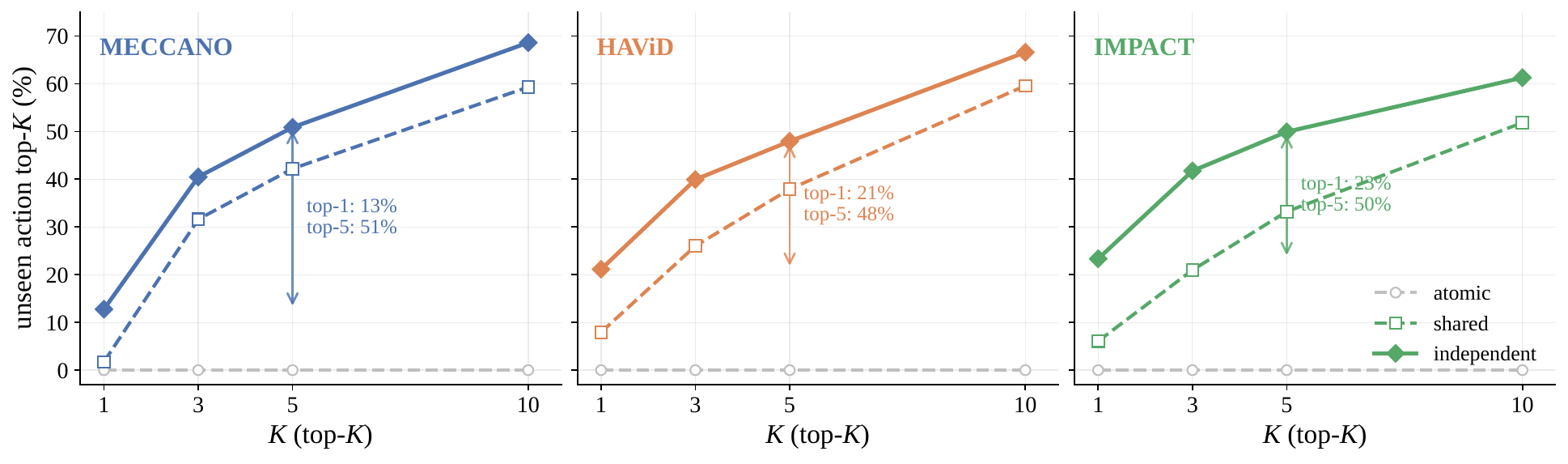}
\caption{
\textbf{Remaining headroom is largely a ranking problem.}
Unseen action top-$K$ under the same scoring function used in Table~\ref{tab:master}. Atomic prediction remains at zero for all $K$, whereas decomposition frequently ranks the correct unseen composition within its top candidates despite low top-1 accuracy.
}
\label{fig:topk}
\end{figure}

Crucially, none of the interventions change the asymmetry identified in RQ2. The larger-vocabulary component remains the dominant source of unseen error across architectures and objectives. Performance improves, but the bottleneck remains. A complementary perspective comes from ranking analysis. Figure~\ref{fig:topk} shows that decomposition frequently ranks the correct unseen composition among its top candidates even when top-1 accuracy is low, whereas atomic prediction remains at zero for all $K$. Much of the remaining error therefore arises from ranking rather than representation: the correct composition is often available but not selected as the top prediction. Taken together, these results suggest that standard interventions improve overall performance without removing the underlying limitations identified in RQ1--RQ3. The remaining headroom lies primarily in ranking unseen compositions correctly rather than in making them reachable.

\section{Discussion: Implications for Compositional Recognition}
\label{sec:discussion}


The four research questions lead to the same conclusion: decomposition improves
reachability but not generalization. Although factorization enables unseen
compositions, performance remains shaped by training co-occurrence structure.
Accuracy is highest in well-supported regions, errors concentrate on the
larger-vocabulary component, and shared training amplifies these effects through
component entanglement. Thus, current decomposition methods behave more like
structured retrieval than genuine compositional reasoning.


This diagnosis suggests three practical directions. First, the dominant
bottleneck can often be anticipated from vocabulary statistics, motivating
asymmetric allocation of model capacity, data, and augmentation. Second,
architecture matters more than objective design: preserving component
separation yields larger gains than adding compositional objectives to
entangled representations. Third, the top-$K$ analysis indicates that much of the remaining error is due to ranking rather than representation, as the
correct unseen composition is often highly ranked but not top-1. Consistent
with this view, subtracting the factorized training prior at inference, without retraining, improves unseen accuracy by up to $+4.2$pp on IMPACT
(Supp.\ Sec.~16). More broadly, our diagnostic framework applies to recognition settings that assume primitive factorization. Published EPIC-KITCHENS-100 baselines show the same vocabulary-asymmetry pattern at roughly ten times our vocabulary scale, with the larger-vocabulary noun component trailing by $14$--$19$pp (Supp.\ Sec.~17). Once reachability is largely addressed, the remaining challenge is to rank unseen compositions correctly.

\section{Conclusion}
\label{sec:conclusion}

Verb--noun decomposition makes unseen compositions reachable, but not
necessarily generalizable. Across three assembly datasets, we find that
compositional performance is largely governed by training-data structure:
primitive support and vocabulary asymmetry determine where failures arise,
while shared training further amplifies them through component entanglement. These results suggest that current decomposition methods rely more on
structured reuse of observed co-occurrences than on genuine compositional
reasoning. We therefore shift the question from whether unseen compositions
are reachable to why they remain difficult once reachable. Our diagnostic
framework links primitive support, vocabulary asymmetry, and component
entanglement to these failures, and we release the associated diagnostics and
the IMPACT compositional split to support future studies of compositional
generalization under primitive factorization.

\section*{Acknowledgements}
This work was funded by Business Finland (grant No 69/31/2025) under the project Human-centric AI-enabled Collaborative Fieldwork Operations (AiWo). The calculations were performed using computational resources provided by the Aalto University Science-IT project.

\bibliography{references}

\end{document}


\maketitle

\noindent
This supplement provides per-(view, hand) and per-subtask breakdowns of the headline results, the
shortcut diagnostic at full per-component resolution, the failure-mode decomposition, calibration
quality with bootstrap confidence intervals, baselines, and the C2C objective ablation. Numbering
follows S$n$/SF$n$ as cited in the main paper. All numbers are produced under the protocol stated in
the main paper, on the same checkpoints; the per-(view, hand) and per-subtask averages reproduce the
aggregate numbers reported in the main tables exactly. Sections~13--17, added following the
reviewers' suggestions, report significance tests for the support correlation, reconstructed
MECCANO held-out splits, a matched-capacity control, a training-prior debiasing intervention, and
an external consistency check on EPIC-KITCHENS-100.

\section{Implementation details}

This section consolidates the experimental setup detail that the main paper
defers here: byte-level training hyperparameters, the frame-sampling protocol,
and the construction details of the compositional splits.

\paragraph{Training hyperparameters.}
Each segment is decoded into $T{=}16$ RGB frames, resized to a short side of
$256$, cropped to $224^2$ (random crop with horizontal flip at training,
centre crop at test), and normalized with per-channel mean $0.45$ and standard
deviation $0.225$. IMPACT's long videos are read by seeking to each segment's
frame range. We optimize with AdamW at base learning rate
$9.6\times10^{-3}$ scaled linearly as $\eta=\eta_\text{base}\,B/256$ with
batch size $B{=}4$, under cosine decay after linear warmup, weight decay
$0.05$, gradient-norm clip $5.0$, dropout $0.5$ before each head, and
cross-entropy with label smoothing $0.1$. Warmup runs for $10$ epochs on
MECCANO and IMPACT and $20$ epochs on HAViD. All single-seed numbers use seed
$42$; the seed-variance table reported later in this supplement (in the C2C
ablation section) reports the $n{=}3$-seed dispersion on MECCANO shared,
which falls well below every effect reported in the main paper.

\paragraph{Frame sampling within each segment.}
At training, the segment is partitioned into $16$ equal-length sub-windows and
one frame per window is drawn uniformly at random. At test, the $16$ frames
are taken at evenly spaced positions across the segment. Segments shorter
than $16$ frames are padded by re-sampling. The sub-window-with-jitter
schedule at training keeps coverage of the segment uniform across epochs while
acting as a mild temporal augmentation.

\paragraph{Compositional split construction.}
The held-out (verb, object) pairs of each compositional split are chosen
manually under fixed quantitative constraints. On HAViD, $13$ (verb,
manipulated-object) pairs are held out, each occurring at least $15$ times in
the dataset and each leaving its verb with at least $5$ and its object with
at least $3$ remaining training partners (about $8\%$ of non-null segments).
On IMPACT, $20$ of $134$ valid pairs per hand are held out ($14.9\%$), each
with at least $10$ segments. On MECCANO, $6$ (verb, object) pairs are held
out under the same neighbour-coverage rule ($462$ unseen segments).
Table~\ref{tab:s1} gives the IMPACT compositional and standard
splits in segment-level detail. The exact held-out pair lists are released
with the code, fixed once before any model is trained, and unchanged across
all reported experiments. No recording crosses the train/test boundary.

\begin{table}[t]
\centering
\caption{\textbf{IMPACT splits constructed here.} Standard is recording-level,
hand-stratified; the compositional split hides $20$ of $134$ pairs per hand
from training. Counts are \textit{front}-view non-null segments.}
\label{tab:s1}
\small\setlength{\tabcolsep}{4pt}
\begin{tabular}{l l c c c}
\toprule
\textbf{Split} & \textbf{Hand} & \textbf{\#rec.} & \textbf{\#seg.} & \textbf{pairs seen / unseen} \\
\midrule
Standard (tr/va/te) & left  & 78/16/18 & 3394/786/849 & 134 / -- \\
Standard (tr/va/te) & right & 75/19/18 & 4719/1197/1086 & 134 / -- \\
\midrule
Comp.\ tr / seen / unseen & left  & -- & 2920/1416/693  & 114 / 20 \\
Comp.\ tr / seen / unseen & right & -- & 3906/1867/1229 & 114 / 20 \\
\bottomrule
\end{tabular}
\end{table}

\section{Per-(view, hand) IMPACT compositional breakdown}
For every (view, hand) configuration of IMPACT, Table~\ref{tab:s2} gives seen / unseen action top-1
and HM under the four architectures studied in the main paper. The mean row reproduces the IMPACT
block of the main-paper anchor table exactly; the per-view dispersion shows the egocentric view
consistently posting the weakest harmonic mean, supporting the deployment caution discussed in the
main paper.
\begin{table*}[t]\centering
\caption{\textbf{IMPACT per-(view,hand) compositional breakdown.} Seen / Unseen action top-1 and HM (\%)
for the five configurations across all 10 (view,hand) configs of IMPACT. Atomic's Un and HM columns are structurally $0$ (the output space is the seen-action vocabulary); the per-view variation in atomic Sn (around $46\!-\!58\%$) is shown for completeness. The mean row matches the IMPACT block of the main-paper anchor; per-view dispersion shows that the egocentric configurations are consistently the weakest.}\label{tab:s2}\scriptsize\setlength{\tabcolsep}{3pt}
\begin{tabular}{l ccc ccc ccc ccc ccc }
\toprule
& \multicolumn{3}{c}{Atomic} & \multicolumn{3}{c}{Shared} & \multicolumn{3}{c}{Independent} & \multicolumn{3}{c}{C2C-shared} & \multicolumn{3}{c}{C2C-indep} \\
\cmidrule(lr){2-4} \cmidrule(lr){5-7} \cmidrule(lr){8-10} \cmidrule(lr){11-13} \cmidrule(lr){14-16}
\textbf{view-hand} & \textbf{Sn} & \textbf{Un} & \textbf{HM} & \textbf{Sn} & \textbf{Un} & \textbf{HM} & \textbf{Sn} & \textbf{Un} & \textbf{HM} & \textbf{Sn} & \textbf{Un} & \textbf{HM} & \textbf{Sn} & \textbf{Un} & \textbf{HM} \\
\midrule
front-lh & 53.1 & 0.0 & 0.0 & 50.6 & 10.8 & 17.8 & 48.9 & 31.6 & 38.4 & 50.7 & 17.9 & 26.5 & 51.6 & 30.3 & 38.2 \\
front-rh & 53.5 & 0.0 & 0.0 & 50.7 & 5.3 & 9.6 & 50.8 & 20.2 & 28.9 & 49.7 & 11.6 & 18.9 & 51.5 & 19.1 & 27.9 \\
left-lh & 54.0 & 0.0 & 0.0 & 56.1 & 8.1 & 14.1 & 55.4 & 27.1 & 36.4 & 41.4 & 17.7 & 24.8 & 53.7 & 26.3 & 35.3 \\
left-rh & 53.9 & 0.0 & 0.0 & 53.0 & 3.3 & 6.1 & 47.1 & 24.7 & 32.4 & 48.0 & 15.8 & 23.8 & 49.5 & 23.4 & 31.7 \\
right-lh & 55.3 & 0.0 & 0.0 & 50.6 & 8.8 & 15.0 & 46.3 & 29.7 & 36.2 & 52.9 & 19.9 & 28.9 & 47.1 & 28.4 & 35.5 \\
right-rh & 53.0 & 0.0 & 0.0 & 51.0 & 4.1 & 7.5 & 47.9 & 18.3 & 26.5 & 50.1 & 15.4 & 23.5 & 50.2 & 20.7 & 29.3 \\
top-lh & 57.5 & 0.0 & 0.0 & 54.5 & 10.4 & 17.5 & 55.6 & 30.4 & 39.3 & 50.4 & 19.9 & 28.5 & 53.5 & 31.2 & 39.4 \\
top-rh & 55.7 & 0.0 & 0.0 & 55.3 & 3.6 & 6.7 & 51.5 & 22.5 & 31.3 & 52.8 & 11.7 & 19.2 & 52.3 & 20.8 & 29.8 \\
ego-lh & 46.0 & 0.0 & 0.0 & 43.8 & 5.2 & 9.3 & 42.7 & 17.3 & 24.6 & 40.4 & 12.6 & 19.3 & 42.2 & 17.0 & 24.2 \\
ego-rh & 46.7 & 0.0 & 0.0 & 44.8 & 1.3 & 2.6 & 43.7 & 11.5 & 18.2 & 43.2 & 4.2 & 7.6 & 43.0 & 11.4 & 18.0 \\
\midrule \textbf{mean} & 52.9 & 0.0 & 0.0 & 51.0 & 6.1 & 10.6 & 49.0 & 23.3 & 31.2 & 47.9 & 14.7 & 22.1 & 49.5 & 22.9 & 30.9 \\
\bottomrule
\end{tabular}\end{table*}

\section{HAViD per-subtask granularity breakdown}
Per-subtask resolution of the granularity sweep is given in Table~\ref{tab:s3}: seen / unseen joint
top-1 and HM for each of the six (view, hand) subtasks across $k\!=\!1\!\ldots\!4$ and the two
single-component solo heads. Every subtask peaks at $k\!=\!2$, which confirms the sweet spot
identified by the aggregate Pareto.
\begin{table*}[t]\centering
\caption{\textbf{HAViD per-subtask granularity breakdown.} Seen / Unseen joint top-1 and HM (\%)
across all 6 (view,hand) subtasks for granularity $k{=}1\!\ldots\!4$ and single-component solo heads.
Substantiates T5 and F4; $k{=}2$ is the per-subtask unseen sweet spot (I5).}\label{tab:s3}\scriptsize\setlength{\tabcolsep}{3pt}
\begin{tabular}{l *{18}{c}}
\toprule
& \multicolumn{3}{c}{$k{=}1$} & \multicolumn{3}{c}{$k{=}2$} & \multicolumn{3}{c}{$k{=}3$} & \multicolumn{3}{c}{$k{=}4$} & \multicolumn{3}{c}{solo $v$} & \multicolumn{3}{c}{solo $m$} \\
\cmidrule(lr){2-4} \cmidrule(lr){5-7} \cmidrule(lr){8-10} \cmidrule(lr){11-13} \cmidrule(lr){14-16} \cmidrule(lr){17-19}
\textbf{subtask} & \textbf{Sn} & \textbf{Un} & \textbf{HM} & \textbf{Sn} & \textbf{Un} & \textbf{HM} & \textbf{Sn} & \textbf{Un} & \textbf{HM} & \textbf{Sn} & \textbf{Un} & \textbf{HM} & \textbf{Sn} & \textbf{Un} & \textbf{HM} & \textbf{Sn} & \textbf{Un} & \textbf{HM} \\
\midrule
v0-lh & 55.9 & 0.0 & 0.0 & 52.0 & 8.3 & 14.3 & 52.0 & 7.8 & 13.6 & 54.6 & 6.0 & 10.8 & 74.9 & 58.2 & 65.5 & 60.3 & 26.3 & 36.7 \\
v0-rh & 50.0 & 0.0 & 0.0 & 48.3 & 6.2 & 11.1 & 44.6 & 6.4 & 11.2 & 50.1 & 5.7 & 10.2 & 71.2 & 59.4 & 64.7 & 55.2 & 28.6 & 37.7 \\
v1-lh & 58.7 & 0.0 & 0.0 & 53.5 & 7.3 & 12.9 & 55.3 & 5.9 & 10.6 & 57.1 & 3.4 & 6.4 & 75.4 & 58.4 & 65.8 & 61.7 & 32.4 & 42.5 \\
v1-rh & 51.3 & 0.0 & 0.0 & 49.5 & 7.1 & 12.4 & 47.3 & 6.2 & 11.0 & 51.8 & 5.4 & 9.8 & 71.5 & 54.8 & 62.0 & 55.4 & 30.5 & 39.4 \\
v2-lh & 56.4 & 0.0 & 0.0 & 53.0 & 9.4 & 16.0 & 52.5 & 9.6 & 16.3 & 57.4 & 7.0 & 12.4 & 73.8 & 57.3 & 64.5 & 60.6 & 37.3 & 46.2 \\
v2-rh & 55.8 & 0.0 & 0.0 & 50.6 & 9.3 & 15.7 & 49.1 & 8.0 & 13.8 & 56.1 & 2.6 & 5.0 & 72.4 & 61.4 & 66.5 & 60.8 & 33.8 & 43.5 \\
\midrule \textbf{mean} & 54.7 & 0.0 & 0.0 & 51.1 & 7.9 & 13.7 & 50.1 & 7.3 & 12.7 & 54.5 & 5.0 & 9.1 & 73.2 & 58.2 & 64.8 & 59.0 & 31.5 & 41.0 \\
\bottomrule
\end{tabular}\end{table*}

\section{Component top-$\{1,3,5\}$ breakdown}
Action top-$\{1,3,5\}$ on the shared and independent variants of all three datasets, the numeric
backing for the main-paper top-$K$ figure, is laid out in Table~\ref{tab:s4}, together with
per-component verb / object top-$\{1,3,5\}$ for the shared encoder where the metric is defined. The
HAViD action top-$K$ values use the same $\log P(v)\!+\!\log P(m)$ scoring as MECCANO/IMPACT and
reproduce the locked top-1 of the main paper exactly.
\begin{table}[t]\centering
\caption{\textbf{Component top-$\{1,3,5\}$ breakdown.} (a) Unseen action composition is
reachable in the top ranks though top-1 is low (numeric backing for F6); atomic is structurally 0 for all
$K$ (omitted). Action top-$K$ uses the same argmax over $(v,n)$ pairs ranked by
$\log P(v)\!+\!\log P(n)$ as the main-paper anchor table; the $K{=}1$ column reproduces the unseen
action top-1 of that anchor exactly. (b) Per-component verb / object top-$K$ (shared, where available).
\emph{Caveat:} MECCANO object uses a coverage metric (no object top-1, omitted).}\label{tab:s4}\small\setlength{\tabcolsep}{4pt}
\begin{tabular}{l l ccc ccc}
\toprule
& & \multicolumn{3}{c}{Seen action top-$K$} & \multicolumn{3}{c}{Unseen action top-$K$} \\
\cmidrule(lr){3-5}\cmidrule(lr){6-8}
\textbf{Dataset} & \textbf{Config} & $K{=}1$ & $3$ & $5$ & $K{=}1$ & $3$ & $5$ \\
\midrule
MECCANO & Shared & 48.4 & 60.1 & 68.6 & 1.7 & 31.6 & 42.2 \\
MECCANO & Independent & 44.3 & 60.2 & 69.3 & 12.8 & 40.5 & 50.9 \\
IMPACT & Shared & 51.0 & 67.5 & 73.1 & 6.1 & 21.0 & 33.2 \\
IMPACT & Independent & 49.0 & 65.6 & 71.5 & 23.3 & 41.7 & 49.9 \\
HAViD & Shared & 51.1 & 68.2 & 73.4 & 7.9 & 26.0 & 37.9 \\
HAViD & Independent & 52.8 & 66.6 & 70.7 & 21.1 & 39.9 & 47.9 \\
\midrule
\multicolumn{8}{l}{\footnotesize (b) per-component top-$K$ (shared encoder):}\\
IMPACT & verb & 63.7 & 82.2 & 88.2 & 26.4 & 47.2 & 59.2 \\
IMPACT & object & 67.1 & 85.8 & 91.3 & 40.9 & 67.9 & 79.8 \\
MECCANO & verb & 63.8 & 87.5 & 93.5 & 39.6 & 86.1 & 91.8 \\
\bottomrule
\multicolumn{8}{l}{\footnotesize Columns under (b) are verb / object top-$K$ for the named component.}\\
\end{tabular}\end{table}

\section{Per-component shortcut diagnostic: solo versus joint}
Diagnostic D1 at full per-component resolution appears in Table~\ref{tab:s5}: the seen-to-unseen
drop for the jointly trained head against the solo head, the excess drop $\Delta_\text{ent}$
(joint minus solo), and the seen-margin (joint $-$ solo on seen). The seen-margin stays close to
zero on every component while $\Delta_\text{ent}$ is large and positive, ruling out capacity as the
cause of the shared encoder's shortfall.
\begin{table}[t]\centering
\caption{\textbf{D1 shortcut diagnostic (solo vs joint).} Per component: seen$\to$unseen accuracy
drop when the head is trained jointly vs in isolation (solo). The \emph{shortcut} $\Delta$ (excess joint
drop) measures reliance on co-occurrence; the seen-margin (joint$-$solo on seen) $\approx 0$ rules out a
capacity deficit. Substantiates T3.}\label{tab:s5}\small\setlength{\tabcolsep}{5pt}
\begin{tabular}{l l cc c c}
\toprule
\textbf{Dataset} & \textbf{Comp.} & joint drop & solo drop & shortcut $\Delta$ & seen-margin \\
\midrule
\multirow{2}{*}{MECCANO} & verb & 24.2 & 17.8 & \textbf{6.4} & -0.3 \\
 & object & 21.7 & 10.2 & \textbf{11.5} & -0.6 \\
\midrule
\multirow{2}{*}{HAViD} & verb & 20.8 & 14.9 & \textbf{5.9} & -1.2 \\
 & manip & 36.2 & 27.5 & \textbf{8.7} & -1.9 \\
\midrule
\multirow{2}{*}{IMPACT} & verb & 37.2 & 24.7 & \textbf{12.5} & +1.8 \\
 & object & 26.2 & 11.7 & \textbf{14.5} & +0.7 \\
\bottomrule
\multicolumn{6}{l}{\footnotesize All in pp. shortcut $\Delta=$ joint drop $-$ solo drop. seen-margin $=$ joint$-$solo seen acc.}\\
\end{tabular}\end{table}

\section{Failure-mode decomposition of unseen errors}
\begin{figure}[t]
\centering
\includegraphics[width=0.8\linewidth]{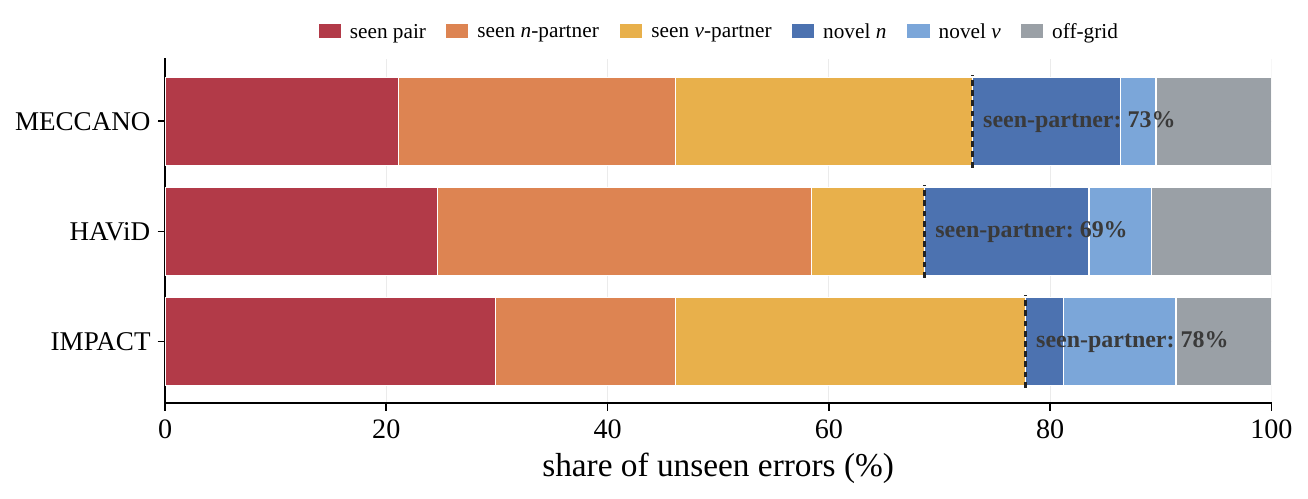}
\caption{\textbf{Where unseen errors land.} Stacked horizontal bars showing the share of wrong
predictions per failure-mode category, per dataset. The dashed line marks the seen-partner total
($73/69/78\%$ on MECCANO/HAViD/IMPACT); the cool segments to its right are predictions that genuinely
leave the grid (off-grid or novel partner).}
\label{fig:sf1}
\end{figure}

Each wrong unseen prediction is attributed to the kind of fallback the model makes in
Table~\ref{tab:s6} and Figure~\ref{fig:sf1}. Across the three datasets, $69$ to $78\%$ of wrong
predictions name a verb or object that was a seen training partner, against only $9$ to $11\%$ that
leave the grid entirely. This is the direct evidence for the co-occurrence-lookup mechanism named
in the discussion of the main text.
\begin{table}[t]\centering
\caption{\textbf{Failure-mode decomposition of unseen errors} (independent, \% of wrong predictions).
The vast majority of errors predict a composition that was a seen training partner/pair (summed in
``seen total''); only a small fraction land off-grid. Direct evidence for I7 (co-occurrence-lookup mechanism).}\label{tab:s6}\scriptsize\setlength{\tabcolsep}{3.5pt}
\begin{tabular}{l cccccc cc}
\toprule
\textbf{Dataset} & seen pair & seen $n$-partner & seen $v$-partner & novel $n$ & novel $v$ & off-grid & \textbf{seen tot.} & off-grid \\
\midrule
MECCANO & 21.1 & 25.1 & 26.8 & 13.4 & 3.2 & 10.4 & \textbf{73} & 10 \\
HAViD & 24.6 & 33.8 & 10.2 & 14.9 & 5.7 & 10.8 & \textbf{69} & 11 \\
IMPACT & 29.9 & 16.3 & 31.6 & 3.5 & 10.2 & 8.6 & \textbf{78} & 9 \\
\bottomrule
\multicolumn{9}{l}{\footnotesize seen $n$/$v$-partner: one component correct, the predicted other was a seen partner. novel: predicted other never co-occurred.}\\
\end{tabular}\end{table}

\section{Calibration and bootstrap confidence intervals}
\begin{figure}[t]
\centering
\includegraphics[width=0.8\linewidth]{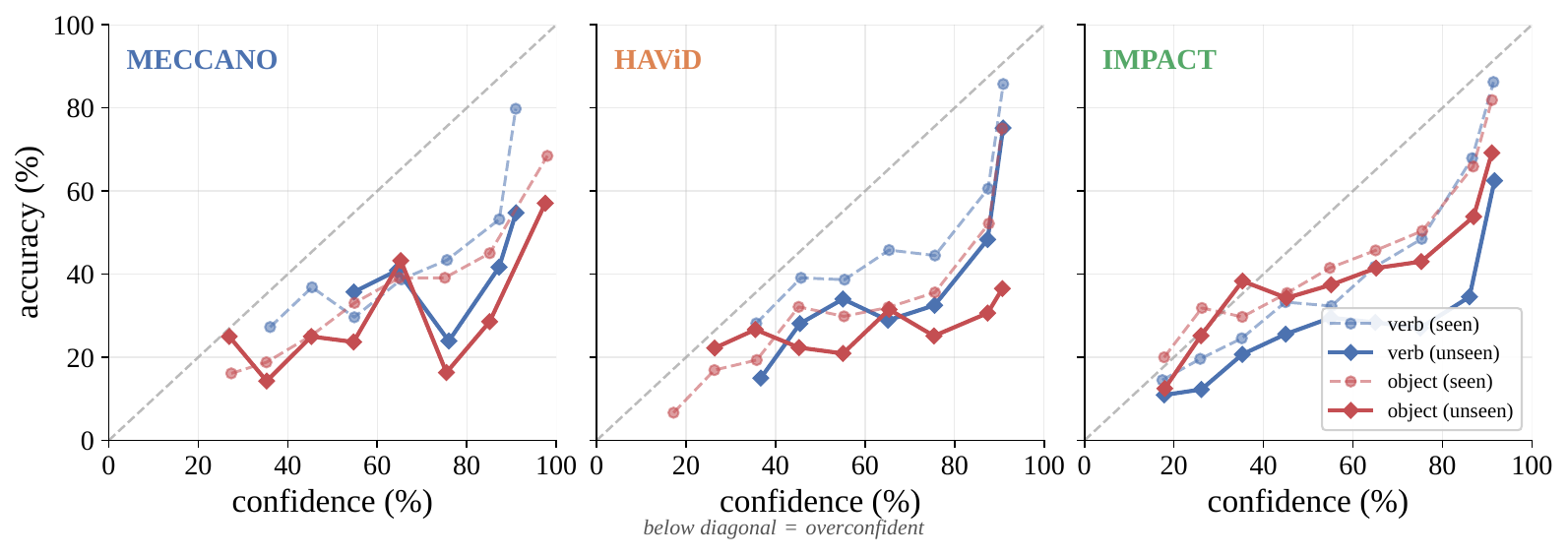}
\caption{\textbf{Calibration reliability of independent recombination.} Per-bin accuracy against
predicted confidence for the verb (blue) and object (red) heads, on seen (dashed) and unseen (solid)
splits. Unseen curves sit below the diagonal: decomposed models are overconfident on novel
compositions, with the gap to the diagonal largest on the bottleneck component (object on
MECCANO/HAViD, verb on IMPACT).}
\label{fig:sf2}
\end{figure}

Expected calibration error (ECE) for the verb and object heads on each dataset, on both seen and
unseen splits, is reported in Table~\ref{tab:s7} alongside $1000\times$ bootstrap $95\%$ confidence
intervals on action top-1. ECE rises from seen to unseen on every dataset and is worst on the
bottleneck component (HAViD object: $0.25 \to 0.50$); the bootstrap intervals confirm that the
seen $\gg$ unseen gap is not sampling noise. Figure~\ref{fig:sf2} shows the corresponding reliability
diagrams.
\begin{table}[t]\centering
\caption{\textbf{Calibration (ECE) and bootstrap 95\% CIs} (independent). ECE rises seen$\to$unseen
for both heads and is worst on the bottleneck component (I6); $1000\times$ bootstrap CI widths on
action top-1 (recentered on the unified argmax-pair top-1 of the main-paper anchor) confirm the
seen$\gg$unseen gap is not sampling noise.}\label{tab:s7}\small\setlength{\tabcolsep}{4pt}
\begin{tabular}{l cc cc cc}
\toprule
& \multicolumn{2}{c}{verb ECE} & \multicolumn{2}{c}{object ECE} & \multicolumn{2}{c}{action top-1 (95\% CI)} \\
\cmidrule(lr){2-3}\cmidrule(lr){4-5}\cmidrule(lr){6-7}
\textbf{Dataset} & seen & unseen & seen & unseen & seen & unseen \\
\midrule
MECCANO & 0.21 & 0.40 & 0.29 & 0.38 & 44.3 (42.8,45.9) & 12.8 (9.7,16.0) \\
HAViD & 0.13 & 0.27 & 0.25 & 0.50 & 52.8 (50.6,55.1) & 21.1 (18.4,24.0) \\
IMPACT & 0.15 & 0.34 & 0.15 & 0.24 & 49.0 (46.6,51.5) & 23.3 (20.7,26.0) \\
\bottomrule
\end{tabular}\end{table}

\section{Unseen baselines}
A majority-class baseline against the independent model's unseen top-1 is contrasted in
Table~\ref{tab:s8}. The majority baseline is $0$ on every dataset by construction, since no seen
action is correct on a held-out composition, so the independent number gives the increment
attributable to decomposition.
\begin{table}[t]\centering
\caption{\textbf{Unseen baselines.} Majority-class prediction collapses to 0 on held-out
compositions (no seen action is correct by construction); the independent model is well above this floor.}\label{tab:s8}\small\setlength{\tabcolsep}{6pt}
\begin{tabular}{l c cc}
\toprule
\textbf{Dataset} & $n_{\text{unseen}}$ & majority top-1 & independent top-1 \\
\midrule
MECCANO & 462 & 0.0 & 12.8 \\
HAViD & 770 & 0.0 & 21.1 \\
IMPACT & 959 & 0.0 & 23.3 \\
\bottomrule
\end{tabular}\end{table}

\section{C2C ablation and seed variance}
\begin{figure}[t]
\centering
\includegraphics[width=\linewidth]{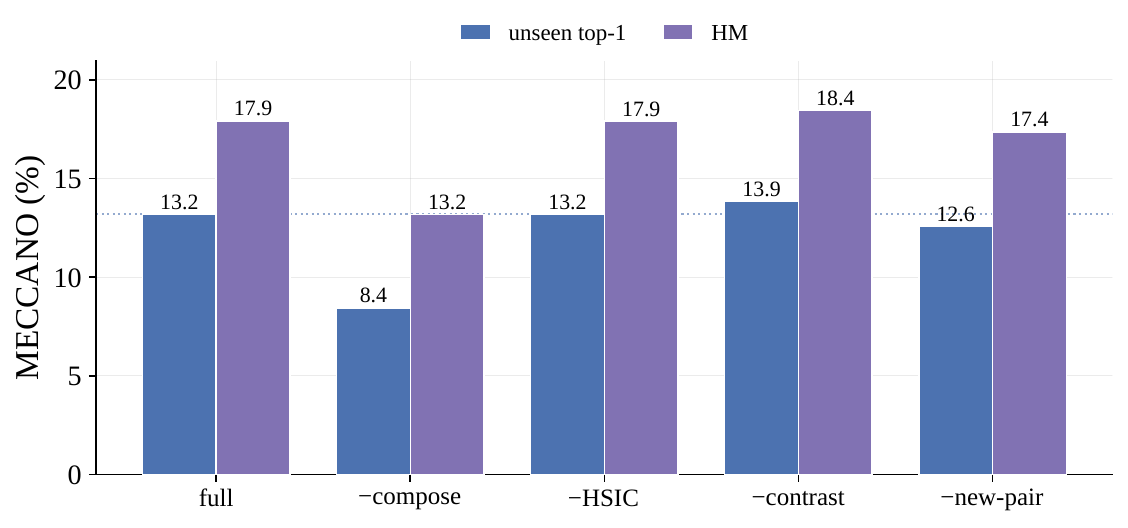}
\caption{\textbf{C2C objective ablation on MECCANO.} Unseen top-1 and harmonic mean under the full
C2C objective and after removing each loss term in turn (composition, HSIC, contrastive, new-pair).
The composition term is the key driver of the C2C gain; removing it cuts unseen top-1 from $13.2$ to
$8.4$. The other terms are near-neutral.}
\label{fig:sf3}
\end{figure}

Two MECCANO ablations are collected in Table~\ref{tab:s9}: (a) removing each of the C2C loss terms
in turn from the full C2C-shared model, and (b) the dispersion of the shared model across three
training seeds. Figure~\ref{fig:sf3} plots panel (a) visually: the composition term drives the C2C
gain, while the HSIC and contrastive terms are near-neutral; the seed dispersion of the shared
baseline ($\pm 0.5$ pp on unseen) is well below the effect sizes reported in the main paper.
\begin{table}[t]\centering
\caption{\textbf{C2C objective ablation (MECCANO) and seed variance.} (a) Removing each compositional
loss term from C2C-shared; (b) MECCANO shared across 3 seeds (mean$\pm$std) confirms the seen$\to$unseen
effects exceed run-to-run noise.}\label{tab:s9}\small\setlength{\tabcolsep}{5pt}
\begin{tabular}{l ccc}
\toprule
\textbf{(a) C2C variant} & Seen & Unseen & HM \\
\midrule
full & 27.9 & 13.2 & 17.9 \\
$-$compose & 29.9 & 8.4 & 13.2 \\
$-$HSIC & 27.8 & 13.2 & 17.9 \\
$-$contrast & 27.5 & 13.9 & 18.4 \\
$-$new-pair & 28.1 & 12.6 & 17.4 \\
\midrule
\multicolumn{4}{l}{\textbf{(b) seed variance}}\\
\midrule
\textbf{config} & Seen & \multicolumn{2}{c}{Unseen} \\
shared ($n{=}3$ seeds) & 48.6 $\pm$ 0.1 & \multicolumn{2}{c}{2.3 $\pm$ 0.5} \\
\bottomrule
\end{tabular}\end{table}

\section{Standard-split breakdown: atomic versus decomposed}
The main-paper anchor reports only the compositional split, since standard-split evaluation has no
seen / unseen concept and is therefore covered separately here. This section gives the standard-split
comparison in three layers: a cross-dataset summary (Table~\ref{tab:s10}), then a per-subtask view of
HAViD (Table~\ref{tab:s11}), and a per-(view, hand) view of IMPACT (Table~\ref{tab:s12}). MECCANO is a
single configuration and so has only a summary row.

The summary in Table~\ref{tab:s10} substantiates the ``masking'' read of the main-paper anchor: under
standard evaluation, decomposition costs little or improves accuracy on every dataset, so a
practitioner choosing between paradigms on a standard benchmark sees no signal of the compositional
gap that the rest of the paper makes visible. HAViD values are taken from the published MViTv2-S
baselines; IMPACT rows average over the $10$ (view, hand) configurations.
\begin{table}[t]\centering
\caption{\textbf{Standard-split breakdown: atomic vs decomposed.} Action top-$\{1,3,5,10\}$ on the standard (non-compositional) test split, with per-component top-1 where the head exists. The atomic$\to$decomposed delta is small or positive on every dataset, which is the ``free lunch'' that masks the compositional question discussed in the main paper. HAViD atomic and decomposed (4-head) top-1 are taken from the published MViTv2-S results~\cite{havid2022} (their 4-head reports per-head top-1 verb $72.4$, manip $59.1$, target $89.7$, tool $98.2$); IMPACT rows average over the $10$ (view, hand) configurations; MECCANO and IMPACT atomic models do not produce per-component scores, and MECCANO's noun metric in the decomposed model is a coverage statistic rather than a top-1.}\label{tab:s10}\small\setlength{\tabcolsep}{4pt}
\begin{tabular}{l l cccc cc}
\toprule
& & \multicolumn{4}{c}{action top-$K$} & \multicolumn{2}{c}{per-component top-1} \\
\cmidrule(lr){3-6}\cmidrule(lr){7-8}
\textbf{Dataset} & \textbf{Config} & $K{=}1$ & $3$ & $5$ & $10$ & verb & object/manip \\
\midrule
MECCANO & Atomic & 47.1 & 67.7 & 74.6 & 83.3 & -- & -- \\
MECCANO & Decomposed & 44.4 & 55.1 & 63.6 & 73.8 & 61.1 & -- \\
IMPACT & Atomic & 48.9 & 65.3 & 71.1 & 77.9 & -- & -- \\
IMPACT & Decomposed & 50.2 & 64.5 & 68.6 & 73.4 & 61.4 & 65.9 \\
HAViD & Atomic & 45.6 & -- & -- & -- & -- & -- \\
HAViD & Decomp.\ (4-head) & 50.9 & -- & -- & -- & 72.4 & 59.1 \\
\bottomrule
\end{tabular}\end{table}

\noindent
Table~\ref{tab:s11} expands HAViD into its six (view, hand) subtasks, reproducing the per-head top-1
of the published 4-head model. Verb and manip carry the bulk of the across-subtask
variance, while target and tool are near-saturated on every subtask. The atomic single-head baseline
is reported only in aggregate by the original work ($45.6\%$ joint top-1).
\begin{table}[t]\centering
\caption{\textbf{HAViD per-subtask standard-split (4-head decomposed).} Per-head top-1 (\%) for the published MViTv2-S 4-head model on each of the six (view, hand) subtasks of HAViD's atomic-action splits. The verb and manip heads carry the bulk of the variance across subtasks; target and tool are near-saturated. The atomic single-head baseline is reported only in aggregate by the original work ($45.6\%$ joint top-1, used in the cross-paradigm summary).}\label{tab:s11}\small\setlength{\tabcolsep}{5pt}
\begin{tabular}{l l ccccc}
\toprule
\textbf{Subtask} & \textbf{tag} & \textbf{Verb} & \textbf{Manip} & \textbf{Target} & \textbf{Tool} & \textbf{Joint} \\
\midrule
Side LH & v0-lh & 74.2 & 58.4 & 90.2 & 98.0 & 51.2 \\
Side RH & v0-rh & 71.2 & 55.1 & 88.9 & 98.2 & 47.0 \\
Front LH & v1-lh & 74.3 & 60.4 & 89.4 & 98.1 & 53.2 \\
Front RH & v1-rh & 69.9 & 55.0 & 88.6 & 98.0 & 46.6 \\
Top LH & v2-lh & 72.5 & 62.8 & 90.5 & 98.7 & 54.2 \\
Top RH & v2-rh & 72.2 & 61.9 & 90.6 & 97.9 & 53.0 \\
\midrule \textbf{mean} & & 72.4 & 58.9 & 89.7 & 98.2 & 50.9 \\
\bottomrule
\end{tabular}\end{table}

\noindent
Table~\ref{tab:s12} gives IMPACT's standard-split breakdown for every (view, hand) configuration,
contrasting the atomic and the decomposed (2-head verb + noun) MViTv2-S models. The mean row matches
the IMPACT entries of Table~\ref{tab:s10}; per-view dispersion is small, and the egocentric configurations
sit a little below the static views even on the standard split.
\begin{table}[t]\centering
\caption{\textbf{IMPACT per-(view, hand) standard-split (atomic vs decomposed).} Action top-$\{1,3,5,10\}$ (\%) for the atomic and decomposed (2-head verb+noun) MViTv2-S models on every (view, hand) configuration of IMPACT, plus per-component top-1 for the decomposed head. The atomic model emits only a flat action distribution, so its verb / noun columns are structurally not defined (``--''). The mean row matches the IMPACT entries of Table~\ref{tab:s10}; per-view dispersion shows that the egocentric configurations sit slightly below the static views even on the standard split.}\label{tab:s12}\scriptsize\setlength{\tabcolsep}{4pt}
\begin{tabular}{l l cccc cc}
\toprule
& & \multicolumn{4}{c}{action top-$K$} & \multicolumn{2}{c}{per-comp.\ top-1} \\
\cmidrule(lr){3-6}\cmidrule(lr){7-8}
\textbf{view-hand} & \textbf{Config} & $K{=}1$ & $3$ & $5$ & $10$ & verb & noun \\
\midrule
front-lh & Atomic & 52.9 & 71.0 & 73.9 & 79.4 & -- & -- \\
front-lh & Decomp & 55.5 & 68.1 & 72.6 & 76.3 & 66.9 & 67.8 \\
front-rh & Atomic & 49.8 & 66.0 & 71.4 & 76.8 & -- & -- \\
front-rh & Decomp & 51.5 & 65.7 & 69.8 & 73.8 & 60.3 & 68.5 \\
left-lh & Atomic & 52.8 & 68.4 & 74.1 & 80.5 & -- & -- \\
left-lh & Decomp & 52.1 & 65.9 & 71.1 & 76.3 & 66.0 & 66.8 \\
left-rh & Atomic & 48.7 & 67.9 & 75.0 & 83.2 & -- & -- \\
left-rh & Decomp & 50.9 & 64.7 & 68.4 & 73.4 & 61.7 & 66.3 \\
right-lh & Atomic & 49.0 & 66.2 & 72.7 & 81.6 & -- & -- \\
right-lh & Decomp & 51.3 & 66.7 & 70.2 & 75.7 & 63.9 & 66.3 \\
right-rh & Atomic & 51.5 & 65.7 & 71.7 & 77.7 & -- & -- \\
right-rh & Decomp & 51.9 & 65.9 & 69.8 & 73.9 & 61.7 & 68.7 \\
top-lh & Atomic & 54.3 & 69.9 & 76.2 & 81.4 & -- & -- \\
top-lh & Decomp & 55.1 & 70.0 & 73.8 & 77.8 & 66.7 & 68.5 \\
top-rh & Atomic & 51.6 & 68.1 & 72.6 & 77.7 & -- & -- \\
top-rh & Decomp & 51.6 & 67.9 & 71.9 & 76.5 & 61.8 & 69.8 \\
ego-lh & Atomic & 41.5 & 55.9 & 63.4 & 72.4 & -- & -- \\
ego-lh & Decomp & 44.4 & 56.6 & 61.6 & 67.5 & 56.4 & 60.2 \\
ego-rh & Atomic & 36.5 & 54.0 & 60.6 & 68.2 & -- & -- \\
ego-rh & Decomp & 37.9 & 53.0 & 57.3 & 63.0 & 48.9 & 56.1 \\
\midrule \textbf{mean} & Atomic & 48.9 & 65.3 & 71.1 & 77.9 & -- & -- \\
\textbf{mean} & Decomp & 50.2 & 64.5 & 68.6 & 73.4 & 61.4 & 65.9 \\
\bottomrule
\end{tabular}\end{table}

\section{Rare versus dense per-pair accuracy}
The coverage-driven reading of the main paper (RQ1 and the discussion) is that accuracy on a pair
tracks the training frequency of that pair rather than whether the pair belongs to the seen or unseen
split. Table~\ref{tab:s13} shows this directly: it bins independent recombination's seen-pair
accuracy by the pair's training frequency and contrasts it with the unseen accuracy on the same
dataset. Densely co-observed pairs ($f{\geq}101$) reach $32$ to $62\%$, while rarely co-observed
seen pairs sit in the low single to low tens of percent, comparable to the unseen accuracy itself.
The seen / unseen partition is therefore not the main axis of difficulty; coverage is.
\begin{table}[t]\centering
\caption{\textbf{Rare versus dense per-pair accuracy.} Per-pair unseen accuracy and per-pair seen accuracy of independent recombination, binned by the pair's training frequency $f$. Cells show mean accuracy in percent (mean $n_\text{pairs}$ in parentheses), averaged over the (view,\,hand) configurations of each dataset; the ``$f{=}1$'' tier is empty on every dataset and is therefore omitted, and the HAViD ``$f{=}2{-}5$'' cell is likewise an empty bin ($n_\text{pairs}{=}0$ on every subtask, since HAViD's seen pairs concentrate in the denser tiers). Densely co-observed pairs ($f{\geq}101$) reach $32$ to $62\%$, while rarely co-observed pairs and unseen pairs sit in the low single to low tens of percent, even though they are technically \emph{seen} in training. This is the coverage-driven pattern referenced in the discussion: accuracy follows training frequency, not whether a pair is in the seen or unseen split.}\label{tab:s13}\small\setlength{\tabcolsep}{5pt}
\begin{tabular}{l ccccc}
\toprule
\textbf{Dataset} & unseen ($f{=}0$) & $f{=}2{-}5$ & $f{=}6{-}20$ & $f{=}21{-}100$ & $f{\geq}101$ \\
\midrule
MECCANO & 11.1\,(6) & 9.1\,(2) & 1.5\,(11) & 18.1\,(21) & 32.3\,(15) \\
HAViD & 18.8\,(13) & -- & 17.8\,(8) & 43.9\,(62) & 60.1\,(19) \\
IMPACT & 19.3\,(20) & 2.6\,(2) & 30.8\,(22) & 41.3\,(37) & 61.9\,(6) \\
\bottomrule
\multicolumn{6}{l}{\footnotesize $f$ is the pair's training-segment count. The ``unseen ($f{=}0$)'' column is the held-out set's accuracy; the remaining columns are seen pairs binned by $f$.}\\
\end{tabular}\end{table}

\section{HAViD C2C intervention, per subtask}
Per-subtask detail for the C2C intervention on HAViD appears in Table~\ref{tab:s14}, covering both
C2C-shared and C2C-indep on each of the six (view, hand) subtasks. The mean row matches the HAViD
C2C rows of the main-paper anchor; the per-subtask dispersion is small, so the aggregate gain
attributed to C2C in the main paper is uniform across subtasks rather than driven by a single one.
\begin{table}[t]\centering
\caption{\textbf{HAViD C2C per-subtask compositional breakdown.} Seen / Unseen action top-1 and HM (\%) of the two C2C variants on each of the six (view, hand) subtasks of HAViD. The mean row matches the HAViD C2C rows of the main-paper anchor; per-subtask C2C-shared dispersion is small (the lift over the plain shared baseline is uniform across subtasks), while C2C-indep tracks the plain independent baseline closely (consistent with the main-paper finding that C2C and de-entanglement are partly substitutive).}\label{tab:s14}\small\setlength{\tabcolsep}{5pt}
\begin{tabular}{l ccc ccc}
\toprule
& \multicolumn{3}{c}{C2C-shared} & \multicolumn{3}{c}{C2C-indep} \\
\cmidrule(lr){2-4}\cmidrule(lr){5-7}
\textbf{subtask} & Sn & Un & HM & Sn & Un & HM \\
\midrule
v0-lh & 54.3 & 19.9 & 29.1 & 54.3 & 17.6 & 26.5 \\
v0-rh & 48.2 & 21.1 & 29.3 & 47.7 & 21.1 & 29.2 \\
v1-lh & 55.7 & 20.2 & 29.7 & 54.0 & 21.3 & 30.6 \\
v1-rh & 50.1 & 17.1 & 25.4 & 48.8 & 20.9 & 29.3 \\
v2-lh & 56.1 & 20.6 & 30.2 & 54.1 & 25.6 & 34.8 \\
v2-rh & 51.7 & 20.8 & 29.7 & 53.2 & 24.8 & 33.9 \\
\midrule \textbf{mean} & 52.7 & 19.9 & 28.9 & 52.0 & 21.9 & 30.7 \\
\bottomrule
\end{tabular}\end{table}

\section{Significance of the support correlation}
\label{sec:supp:significance}
We assess the statistical reliability of the RQ1 support correlation. Per-pair unseen accuracies
are pooled over configurations ($13$ pairs $\times$ $6$ subtasks on HAViD; $20$ pairs per hand
$\times$ $10$ (view, hand) configurations on IMPACT), giving one (support, accuracy) point per
pair--configuration combination. We report the pooled Spearman $\rho$, a permutation test that
shuffles accuracies over pairs, and a bootstrap $95\%$ confidence interval ($1000$ resamples).
Table~\ref{tab:supp:sig} summarizes the outcome: the correlation is highly significant on HAViD
and IMPACT, whereas MECCANO alone is underpowered with only six held-out pairs. Sign consistency
supports the same conclusion at the configuration level: all $17/17$ per-configuration correlations
are positive (sign test $p{=}0.001$ over the ten IMPACT configurations and $p{=}0.016$ over the six
HAViD subtasks).

\begin{table}[h]
\centering
\caption{Significance of the per-pair support correlation (independent recombination).}
\label{tab:supp:sig}
\begin{tabular}{lcccc}
\toprule
Dataset & $n$ (pair--config points) & Spearman $\rho$ & Permutation $p$ & Bootstrap $95\%$ CI \\
\midrule
HAViD   & 78  & 0.70 & $5\times10^{-5}$ & $[0.56,\,0.81]$ \\
IMPACT  & 200 & 0.51 & $5\times10^{-5}$ & $[0.39,\,0.61]$ \\
MECCANO & 6   & 0.49 & 0.17 & -- \\
\bottomrule
\end{tabular}
\end{table}

\section{Reconstructed MECCANO held-out splits}
\label{sec:supp:altsplits}
MECCANO's original held-out set contains only six pairs, all composed of the two highest-frequency
verbs, which both limits statistical power and confines the held-out support range to the top of
the distribution. We therefore rebuilt two alternative held-out splits under the same construction
constraints as the original (every verb and noun retains sufficient partners in training and the
unseen set keeps a comparable segment budget). Each alternative split replaces three of the six
held-out pairs, which is the maximum the dataset's skewed composition graph admits: only the two
dominant verbs offer enough co-occurrence partners to be held out more aggressively. Shared and
independent models are retrained from scratch on each split.

Table~\ref{tab:supp:alt} shows two results. First, the independent-versus-shared unseen gap
reproduces on both alternative splits ($+12.6$ and $+18.4$pp), so the main-paper conclusion does
not depend on the particular held-out set. Second, because the reconstructed held-out pairs span a
wider support range, the per-pair support correlation rises to $\rho{=}0.94$ and $0.93$: MECCANO
exhibits the RQ1 pattern strongly once its held-out set is not confined to the highest-frequency
pairs, confirming that the weak $\rho{=}0.49$ on the original split is an artifact of its narrow
support range rather than a counterexample to RQ1.

\begin{table}[h]
\centering
\caption{Alternative MECCANO held-out splits: seen\,/\,unseen action top-1 (\%) and per-pair
support correlation on the unseen set.}
\label{tab:supp:alt}
\begin{tabular}{lccc}
\toprule
Split & Shared & Independent & $\rho$ (unseen pairs) \\
\midrule
Original    & 48.4\,/\,1.7 & 44.3\,/\,12.8 & 0.49 \\
Alternative 1 & 48.4\,/\,1.7 & 47.8\,/\,14.3 & 0.94 \\
Alternative 2 & 48.6\,/\,2.4 & 44.4\,/\,20.8 & 0.93 \\
\bottomrule
\end{tabular}
\end{table}

\section{Matched-capacity control}
\label{sec:supp:frozen}
Independent recombination uses one backbone per component, so its advantage over the shared
formulation could in principle stem from the larger trainable-parameter count rather than from
component separation. To control for capacity we train a frozen-half independent variant: in each
tower we freeze the patch-embedding projection, the positional encodings, and the first $12$ of
$16$ MViTv2-S blocks, leaving ${\sim}17.2$M trainable parameters per tower, i.e.\ $34.4$M in total,
which matches the shared model's $34.2$M. All other training settings are unchanged.

As Table~\ref{tab:supp:frz} shows, the frozen-half variant still reaches $8.4\%$ unseen top-1 on
MECCANO ($5\times$ the shared model's $1.7\%$) and $20.4\%$ on the IMPACT front view ($2.5\times$
the shared model's $8.1\%$; mean over the two front-view hand configurations). Freezing lowers seen
accuracy as expected, so the unseen advantage cannot stem from greater fitting capacity: component
separation, not trainable-parameter count, drives the shared--independent gap of the main paper.

\begin{table}[h]
\centering
\caption{Matched-capacity control: seen\,/\,unseen action top-1 (\%). Frz$\tfrac{1}{2}$ freezes
half of each independent tower so that total trainable capacity matches the shared model.}
\label{tab:supp:frz}
\begin{tabular}{lccc}
\toprule
 & Shared & Independent & Independent-frz$\tfrac{1}{2}$ \\
\midrule
MECCANO & 48.4\,/\,1.7 & 44.3\,/\,12.8 & 45.8\,/\,8.4 \\
IMPACT (front, 2 configs) & 50.7\,/\,8.1 & 49.9\,/\,25.9 & 45.4\,/\,20.4 \\
\bottomrule
\end{tabular}
\end{table}

\section{Training-prior debiasing without retraining}
\label{sec:supp:debias}
The main-paper discussion attributes much of the remaining unseen error to ranking: the correct
composition is often highly ranked but loses to combinations favoured by the training co-occurrence
prior. We test the smallest intervention this analysis suggests: subtracting the factorized
training prior from the recombined scores of the already-trained independent models,
$s'(v,n) = \log P(v) + \log P(n) - \tau\,[\log \hat P(v) + \log \hat P(n)]$, where $\hat P$ are
the training marginals and $\tau$ controls the strength of the removal. No parameter is retrained.
On MECCANO, whose noun target is a set, the noun prior uses per-noun Bernoulli marginals.

Table~\ref{tab:supp:debias} reports the sweep over $\tau$. On IMPACT ($8$ static-view
configurations) unseen top-1 improves from $25.6$ to $29.8$ ($+4.2$pp) at $\tau{=}1$ with the
harmonic mean improving by $+2.6$pp; HAViD gains $+0.8$pp unseen at $\tau{=}0.5$. On MECCANO the
intervention brings no gain ($12.8\to10.8$ at $\tau{=}1$): its six held-out pairs involve the two
highest-frequency verbs, so the unseen mass lies in the high-support region where removing the
prior is not expected to help. The success profile of the intervention therefore follows the
support structure of RQ1, corroborating the diagnosis independently.

\begin{table}[h]
\centering
\caption{Prior debiasing at inference: unseen top-1\,/\,harmonic mean (\%) as a function of the
debiasing strength $\tau$, averaged over configurations.}
\label{tab:supp:debias}
\begin{tabular}{lccccc}
\toprule
 & $\tau{=}0$ & $\tau{=}0.25$ & $\tau{=}0.5$ & $\tau{=}0.75$ & $\tau{=}1$ \\
\midrule
IMPACT (8 configs) & 25.6\,/\,33.7 & 26.3\,/\,34.3 & 27.5\,/\,35.2 & 28.6\,/\,36.0 & \textbf{29.8}\,/\,\textbf{36.2} \\
HAViD (6 subtasks) & 21.1\,/\,30.1 & 21.8\,/\,30.6 & \textbf{21.9}\,/\,\textbf{30.7} & 21.9\,/\,30.5 & 18.3\,/\,26.2 \\
\bottomrule
\end{tabular}
\end{table}

\section{External consistency: EPIC-KITCHENS-100}
\label{sec:supp:epic}
RQ2 predicts that the failing component follows the larger vocabulary. EPIC-KITCHENS-100
(Damen \etal, IJCV 2022) offers a test of this regularity at roughly ten times our vocabulary
scale: it has $97$ verbs and $300$ nouns ($3.1\times$ asymmetry), so the prediction is a noun
bottleneck. Every published baseline is consistent with it: verb\,/\,noun top-1 accuracy is
$60.2\,/\,46.0$ for TSN, $67.9\,/\,49.0$ for TSM, and $65.6\,/\,50.0$ for SlowFast, i.e.\ the
larger-vocabulary component trails by $14$--$19$pp in every case. We note that these are published
standard-split numbers under different backbones and protocols, so they provide consistency at
scale rather than a controlled comparison; the controlled evidence for the vocabulary-asymmetry
regularity remains IMPACT's verb--noun flip in the main paper. Running the full five-formulation
diagnosis on EPIC's compositional splits is left to future work; the released protocol applies
directly.